\pdfoutput=1

\documentclass[dvipsnames, 11pt]{article}

\usepackage[preprint]{acl}

\usepackage{times}
\usepackage{latexsym}

\usepackage[T1]{fontenc}

\usepackage[utf8]{inputenc}

\usepackage{microtype}

\usepackage{inconsolata}

\usepackage{colortbl}
\usepackage{adjustbox}
\usepackage{diagbox}
\usepackage{amsmath,amssymb, amsthm}
\usepackage[inline]{enumitem}
\usepackage{array,multirow,graphicx,makecell}
\usepackage{booktabs}
\usepackage{tabularx}
\usepackage{xfrac}
\usepackage[skins]{tcolorbox}
\tcbuselibrary{breakable}
\usepackage{anyfontsize}
\usepackage{framed}
\usepackage{tabto}
\usepackage{tabulary}
\usepackage{algpseudocode}
\usepackage{algorithm}
\usepackage{listings}
\usepackage{anyfontsize}
\usepackage[toc,page,header]{appendix}
\usepackage{minitoc}

\newcounter{tcbcounter}

\usepackage{amsmath,amsfonts,bm}

\def\eqref#1{equation~\ref{#1}}

\def\1{\bm{1}}

\DeclareMathAlphabet{\mathsfit}{\encodingdefault}{\sfdefault}{m}{sl}
\SetMathAlphabet{\mathsfit}{bold}{\encodingdefault}{\sfdefault}{bx}{n}

\newif\ifcomments
\commentstrue
\ifcomments
    \providecommand\roi[1]{\textcolor{purple}{[Roi: #1]}}    \providecommand\rotem[1]{\textcolor{blue}{[Rotem: #1]}}
    \providecommand\nitay[1]{\textcolor{magenta}{[Nitay: #1]}}
    \providecommand\todo[1]{\textcolor{red}{[TODO: #1]}}
\else
    \providecommand{\roi}[1]{}    \providecommand{\rotem}[1]{}
    \providecommand{\nitay}[1]{}
    \providecommand{\todo}[1]{}
\fi

\newtcolorbox[auto counter, number within=section]{prompt}[3][]{%
  enhanced,
  breakable,
  colback=#2!5!white,
  colframe=#2!75!black,
  title=\textbf{Box \thetcbcounter: #3},
  fontupper=\footnotesize\fontfamily{cmr}\selectfont,
  #1
}

\newtcolorbox{quotebox}{
enhanced,
boxrule=0pt,frame hidden,
borderline west={4pt}{0pt}{CadetBlue!90!black},
colback=CadetBlue!20!white,
sharp corners
}

\title{Multi-Domain Explainability of Preferences}

\author{Nitay Calderon$^T$ \And  Liat Ein-Dor$^I$\\
        $^T$Faculty of Data and Decision Sciences, Technion \quad $^I$IBM Research \\ 
        \texttt{nitay@campus.technion.ac.il} \quad \texttt{liate@il.ibm.com} \quad \texttt{roiri@technion.ac.il}
         \\
        \And Roi Reichart$^T$}

\begin{document}

\maketitle

\doparttoc 
\faketableofcontents 

\begin{abstract}
Preference mechanisms, such as human preference, LLM-as-a-Judge (LaaJ), and reward models, are central to aligning and evaluating large language models (LLMs). Yet, the underlying concepts that drive these preferences remain poorly understood. In this work, we propose a fully automated method for generating local and global concept-based explanations of preferences across multiple domains. Our method utilizes an LLM to identify concepts that distinguish between chosen and rejected responses, and to represent them with concept-based vectors. To model the relationships between concepts and preferences, we propose a white-box Hierarchical Multi-Domain Regression model that captures both domain-general and domain-specific effects. To evaluate our method, we curate a dataset spanning eight challenging and diverse domains and explain twelve mechanisms. Our method achieves strong preference prediction performance, outperforming baselines while also being explainable. Additionally, we assess explanations in two application-driven settings. First, guiding LLM outputs with concepts from LaaJ explanations yields responses that those judges consistently prefer. Second, prompting LaaJs with concepts explaining humans improves their preference predictions. Together, our work establishes a new paradigm for explainability in the era of LLMs.\footnote{\url{https://github.com/nitaytech/PrefExplain}}
\end{abstract}

\section{Introduction}
\label{sec:intro}

\begin{figure}[!t]
    \centering
    \includegraphics[width=0.475\textwidth]{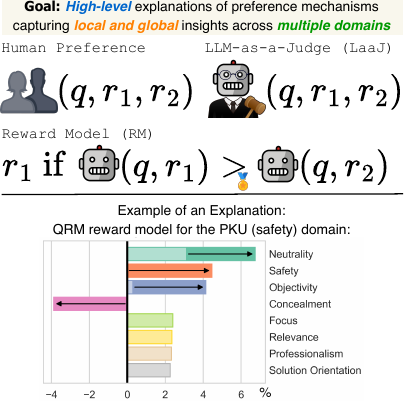}
    \caption{\textbf{Top -- Preference Mechanisms:} Given a triplet of a user query $q$ and two responses, $r_1$ and $r_2$, each mechanism selects a preferred response. Our work automatically discovers concepts for global explanations within a multi-domain learning framework, providing a structured view of how concepts influence preferences generally (shared) and within each domain (specific). \textbf{Bottom -- An Explanation:} Explaining the QRM reward model in the PKU domain. Lighter bars represent the impact of shared concepts, while darker bars and arrows indicate domain-specific deviations.}
    \label{fig:mechanisms}
\vspace{-0.8em}
\end{figure}

\textit{Preference mechanisms}, which select a preferred response to a given user query, are central to \textit{Large Language Model (LLM)} research and form a core component of both alignment and evaluation of LLMs. Three leading preference mechanism types, human preference, LLM-as-a-Judge, and reward models, are illustrated in Figure~\ref{fig:mechanisms}. \textit{Human preference} is widely used to train reward models \citep{Kaufmann23RLHF}, to directly align LLMs through methods like direct preference optimization (DPO) \citep{RafailovSMMEF23}, and remains one of the most reliable approaches for benchmarking LLMs \citep{ChiangZ0ALLZ0JG24}.
The \textit{LLM-as-a-Judge (LaaJ)} paradigm, which employs LLMs to evaluate other LLMs\cite{zheng2023judging}, has become the de facto standard for automatic evaluation \citep{Li24LLMASJUDGE}. LaaJs can even replace humans for alignment, as in RLAIF \citep{PMMFLBHCRP24}.
Finally, \textit{Reward Models (RMs)} are paramount in RLHF and broader LLM alignment efforts \citep{Wang24survey}.

Despite their centrality to LLM research and development, the underlying \textit{concepts} (often referred to as attributes, factors, features, or properties) driving preferences remain inadequately understood. A growing body of work shows that preferences can be influenced by response length \citep{singhal2023towards}, sycophancy \citep{sharma2023towards} and writing style \citep{gudibande2023false}
and that LLMs may favor responses that resemble their own \citep{zheng2023judging}.
\citet{li2024dissecting} study $29$ concepts, showing that humans are sensitive to concepts like politeness and stance alignment, whereas LaaJs prioritize factuality and safety. 

Such carefully curated analyses offer a promising direction. However, they rely on concepts predefined by researchers, potentially biasing the analysis toward preconceived notions. Additionally, they typically require manual annotation, limiting the scalability of the analysis. Finally, existing studies are often constrained to a single domain or dataset, leaving open the question of whether influencing concepts vary across diverse domains. 

In this work, we propose a fully automated method for concept-based explainability of preferences across multiple domains. 
Our method, consisting of four stages, is described in \S\ref{sec:method} and illustrated in Figure~\ref{fig:method}. First, preference data is used by an LLM to discover concepts that differentiate between chosen and rejected responses (\S\ref{sub:discovering}). Next, an LLM is used to represent each triplet of user query and two responses as a concept vector (\S\ref{sub:predicting}). Then, a logistic regression-based model is trained to predict preferences from these vectors. Finally, model weights reveal concept importance, providing meaningful explanations as shown in Figure~\ref{fig:mechanisms}.

A special focus of our work is on a key challenge in preference explainability: preferences are domain-dependent. Concepts influence preference decisions differently across domains, and a concept that is critical in one domain may be less relevant in another. For instance, the \textit{`Concealment'} concept is highly relevant when evaluating responses in safety-focused domains (see Figure~\ref{fig:mechanisms}), but has little relevance in domains centered on food recipes. Accordingly, we propose in \S\ref{sub:model} a \textit{hierarchical multi-domain regression (HMDR)} model, a white-box model that decomposes concept weights into a \textit{shared component}, which is fixed across domains, and a \textit{domain-specific component}, which captures variations unique to each domain. Unlike traditional multi-task approaches, our model is explicitly optimized for domain generalization, requiring the shared component to be predictive on its own while also promoting concept sparsity.

To evaluate our method, we curated eight challenging preference datasets spanning diverse domains (\S\ref{sub:data}). We explain a total of twelve mechanisms (\S\ref{sub:models}), including human preference, two reward models, and nine LLM judges, covering various LLMs and prompting techniques such as chain-of-thought, in-context learning, and LLM ensembles. After conducting a human evaluation and statistically validating the LLM-based concept annotations, we assess the preference prediction performance of our method. 
Our results (\S\ref{sec:results}) show that our explainable method achieves comparable performance to or even better than black-box alternative methods. An ablation study of the HMDR model further demonstrates that this white-box model excels not only in in-domain settings but also in out-of-domain generalization.

To evaluate the quality of our explanations, we introduce two application-driven settings. In the first, \textit{Judge Hack}, we test whether our explanations identify concepts that truly matter to LaaJs and RMs. We prompt LLMs to produce responses conditioned on the top-ranked concepts from a given judge and find that the judge consistently prefers these explanation-guided responses over others. In the second, \textit{Tie Break}, we apply our explanations to resolve tie cases, occurring when LaaJs give inconsistent predictions depending on response position (10\%-30\% of the times). We use explanations to identify the most important concepts to humans and re-prompt the judge to resolve ties based on these concepts. This procedure leads to consistent gains, up to 10\%, in alignment with human preferences. 

To summarize, our contributions are:
(1) We propose an fully automated method for preference explainability: discovering concepts, representing examples as concept vectors, and modeling relationships between concepts and preferences; (2) We introduce the HMDR model, enabling multi-domain learning for explainability; (3) We curate a dataset spanning eight diverse domains and explain twelve mechanisms; (4) We propose two application-driven evaluation settings. Together, our work provides a new paradigm for explainability in the era of LLMs.
\section{Related Work}
\label{sec:related}

We discuss related work here and refer the reader to Appendix \ref{app:back_exp} for background on NLP explainability and concept-based explanations, and to Appendix \ref{app:back_md} for multi-domain learning. 

Concept-based explanations are increasingly favored over token-level methods due to their alignment with human reasoning and ability to support both local and global insights \citep{kim2018tcav, KimWRFM23, GatCFCSR24}. One approach to concept-based explainability is concept bottleneck models \citep{koh2020conceptbottleneck}, which use interpretable concepts as intermediate variables. Recent work leverages LLMs to discover concepts for bottleneck models \citep{ludan2023tbm, sun2024cblm}, though typically in simple classification tasks such as sentiment analysis and topic classification, and does not address a multi-domain setup. Our focus is on concept discovery for preference mechanisms.

Several studies analyze preferences using regression over manually defined concepts \citep{sharma2023towards, HoskingBB24, li2024dissecting}, while others generate counterfactuals to assess causal effects of concepts \citep{jiang2025contrastive}, or train multi-objective reward models that jointly model concepts and preference scores \citep{wang2024multiobjective, qrm}. Another relevant line of work focuses on generation evaluation, scoring generated-text along multiple dimensions, without aggregating them into a final preference \citep{jiang2024tiger, kim2024prometheus, kim2024prometheus2, patronus2024glider}. Unlike all of these works, our method \textit{automatically discovers concepts} from the data and explains multiple PMs in a \textit{multi-domain setting}.

Our HMDR model builds on insights from both domain-invariant \citep{ganin2016domain, ziser2018pivot, arjovsky2019invariant} and domain-specific learning \citep{Ben-DavidOR22, VolkBACR23}. We are also inspired by the Dirty Model from multi-task learning for regression tasks \citep{JalaliRSR10}, which decomposes model weights into shared and task-specific components. Our HMDR model, designed for classification tasks, is optimized for domain generalization and supports different sparsity structures.
\section{Method}
\label{sec:method}
\begin{figure}[t]
    \centering
    \includegraphics[width=0.495\textwidth]{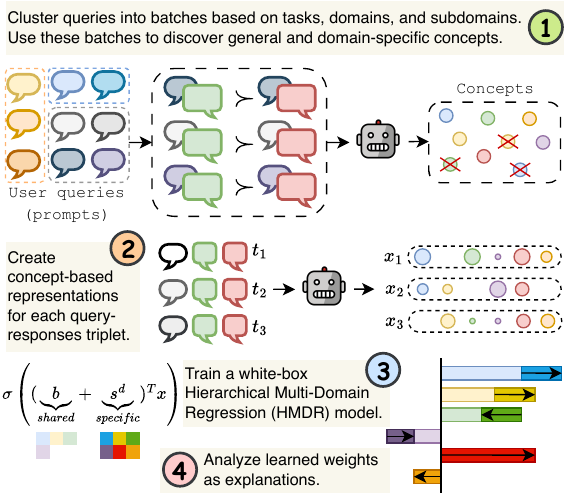}
    \caption{\textbf{Method Illustration:} Given a dataset of triplets $(q, r_1, r_2)$, our four-stage method generates both local and global explanations of preference mechanisms.} 
    \label{fig:method}
\vspace{-0.8em}
\end{figure}

Our method assumes access to preference data collected across multiple domains. Each data point from domain $d$ is a triplet $t^{(d)} = (q, r_1, r_2)$, where $q$ is a \textit{user query}, and $r_1$, $r_2$ are two responses, produced either by humans or LLMs. A preference mechanism assigns one response as \textit{chosen} ($r_{+}$) and the other as \textit{rejected} ($r_{-}$).
\footnote{We use binary labels: 1 if $r_1=r_{+}$, and -1 if $r_2=r_{+}$.} 
Given a mechanism, our method automatically generates a global explanation by generating a set of human-interpretable concepts and finding their relative impact on the decision of the mechanism. 
The method consists of four stages as illustrated in Figure~\ref{fig:method}. We next describe each stage in detail.

\subsection{Concepts Discovery}
\label{sub:discovering}

In the first stage we employ an LLM to propose potential concepts that may explain \textit{why the chosen responses were preferred over the rejected ones.}
Since the concepts underlying the mechanism may vary across domains, this stage is done for each domain separately, allowing the discovery of domain-specific concepts. Note that the domain of each triplet is provided in the data.


\paragraph{Batching User Queries} 
We group triplets from each domain $d$ into batches of size $n_b$, where each batch corresponds to a subdomain or task (e.g., question answering, explanation, summarization, advice). This serves two purposes: (1) batching may encourage discovering concepts less specific to individual instances, and (2) batching by subdomain or task may help identify domain-specific concepts relevant to the batch. To achieve this, we randomly sample 10\% of the queries and prompting an LLM to generate a list of relevant subdomains and tasks for each query. Next, we retain the ten most frequent subdomains and tasks and use the LLM to annotate every query $q \in d$ with subdomain(s) and task(s) from this list. Finally, we construct $B$ batches of size $n_b$ by randomly sampling examples that share either the same subdomain, the same subtask, both, or neither.

\paragraph{Discovering Candidate Concepts} Each batch of triplets, along with its subdomain and task, is provided to the LLM for concept discovery. The LLM is tasked to propose $n_c$ concepts that may explain why one response is preferred over the other. Additionally, we ask the LLM to generate a concise one-sentence description of each proposed concept. To introduce variability, we slightly adjust the prompt randomly for each batch by asking why the first response was chosen over the second, why the second was less favorable, or why a given response was either selected or rejected when presented alone. 

\paragraph{Filtering and Defining Concepts} At this stage, the LLM has identified up to $B \times n_c$ concepts per domain. Many of these are duplicates, either exact or semantically similar (e.g., `relevance to user query' and `relevancy'). 
To filter semantic duplicates, we first apply word stemming using the Snowball stemmer \citep{BirdL04}. Two concepts are flagged as potential duplicates if they share at least one stemmed word (e.g., `relev' in the example above). We then use an LLM to determine whether the flagged concepts are semantically similar. If they are, we retain only the more frequent concept, prioritizing those found in more domains, and in the case of a tie, in more batches. After filtering, we are left with $c$ concepts. We then identify the set of \textit{shared concepts}: concepts discovered in at least half of the domains. Other concepts are termed \textit{specific concepts}. Finally, we ask the LLM to generate concept definitions by conditioning on up to five generated descriptions from the discovery stage using the format: \textit{``A high score indicates the response...; a low score indicates the response...''.}

\subsection{Concept-based Representations}
\label{sub:predicting}

Given the discovered concepts, we aim to represent each triplet as a \textit{concept vector}, where each feature corresponds to a concept. To reduce the number of LLM calls, we first predict the relevant concepts for each user query $q \in d$. Given the shared concepts and the concepts specific to domain $d$, the LLM generates a filtered list containing only those it believes are relevant to $q$. 

We use the relevant concepts to construct the representation $x^{(d)} \in \mathbb{R}^{c}$. Note that irrelevant concepts, including those not specific to $d$, are automatically assigned a value of 0. We propose and explore two types of concept-based representations:
\textbf{Comp-rep:} The entire triplet is provided to the LLM, which predicts a value of 1 if the first response better aligns with the concept definition (i.e., it would be preferred based on the concept alone), -1 if the second response aligns better, and 0 if both are equally aligned.
\textbf{Score-rep:} Each response is scored independently by the LLM on a 1–7 scale based on the concept definition. The final concept value is the difference between the scores of the first and second responses.

\subsection{Hierarchical Multi-Domain Regression}
\label{sub:model}

Given the concept-based representations, we train a white-box model to predict the decisions of a given mechanism. The model weights are then used to explain that mechanism. To support multi-domain explainability, we propose a \textit{mixed-effects model} that captures the hierarchical structure of the data \citep{snijders2011multilevel}. 
While some concepts may have a consistent effect across all domains, others may exhibit domain-specific behavior.

Inspired by the multi-task learning \textit{dirty model} of \citet{JalaliRSR10}, we introduce a \textit{Hierarchical Multi-Domain Regression (HMDR)} model, which decomposes the regression weights into a \textit{shared vector} and \textit{domain-specific deviations}. To further support out-of-domain generalization, we incorporate an additional loss term that explicitly encourages the shared component to be predictive on its own. Finally, we apply regularization terms to promote sparsity in both components, enhancing interpretability \citep{poursabzi2021manipulating}.
We now describe the HMDR model in detail.


For each domain $d\in\{1,\ldots,D\}$ we observe $X^{(d)}\in\mathbb{R}^{n_d\times c}$ and $y^{(d)}\in\{-1,1\}^{n_d}$, where $n_d$ is the number of instances in domain $d$ and $c$ is the number of \textit{concepts}. 
The matrix $X^{(d)}$ contains Comp- or Score-representations, and the labels $y^{(d)}$ may come from humans, LaaJs, or RMs. 

The logistic regression weights of domain $d$ are:
\[
\beta^{(d)} = b + s^{(d)}
\]
where $b\in\mathbb{R}^{c}$ is the \textit{shared weight vector} common to all domains and $s^{(d)}\in\mathbb{R}^{c}$ is the \textit{domain-specific deviation vector}.\footnote{$b$ contains zeros for non-shared concepts. $s^{(d)}$ contains zeros for concepts not specific to $d$, while shared concepts may have non-zero values. }
The predicted probability is:
\[
\hat{p}^{(d)}_i = \sigma\Big({\beta^{(d)}}^\top x^{(d)}_i\Big) =\frac{1}{1+\exp\Big(-{\beta^{(d)}}^\top x^{(d)}_i\Big)}
\]
In case the domain is unknown, we set $s^{(d)}=0$, and the weights are $\beta=b$. The logistic loss is:
\[
\ell\big(y, \beta^\top x\big) = \log\Big(1+\exp\big(-y\, (\beta^\top x)\big)\Big)
\]
We propose the following optimization objective:
\begin{align}
\min_{b,\{s^{(d)}\}_{d=1}^D} \quad & \sum_{d=1}^D \Biggl\{ \underbrace{\sum_{i=1}^{n_d} \ell\Bigl(y^{(d)}_i,(b+s^{(d)})^\top x^{(d)}_i\Bigr)}_{\text{Domain-specific loss}} \; \nonumber\\
& \quad +\; \underbrace{\alpha \sum_{i=1}^{n_d} \ell\Bigl(y^{(d)}_i,\,b^\top x^{(d)}_i\Bigr)}_{\text{Shared loss}} \Biggr\} \\
& \quad + \; \lambda_b \|b\|_1 \;+\; \lambda_s \sum_{d=1}^D \|s^{(d)}\|_1 \nonumber
\label{eq:objective}
\end{align}
The overall objective is a sum over the domains, where each domain contributes both a \textit{domain-specific loss} (using $b+s^{(d)}$) and a \textit{shared loss} (using $b$), together with $\ell_1$ regularizers that promote sparsity. $\alpha\geq 0$ is a hyperparameter that balances the importance of the shared loss relative to the domain-specific losses. The hyperparameters $\lambda_b>0$ and $\lambda_s>0$ control the weight sparsity.

\paragraph{From a Model to an Explanation} In the fourth stage, we analyze our model weights to derive explanations. We quantify the importance of a concept $c_j$ in domain $d$ by measuring the \emph{lift} in predicted probability: $100 \cdot \Delta\hat {p}^{(d)}_j / \hat{p}^{(d)}$. For local explanations, $\Delta\hat {p}^{(d)}_j$ is the difference between the probability of incrementing $c_j$ by one, with all other concepts held fixed, and the probability for the original input. For global explanations, we approximate the expected lift as $50\cdot (b_j + s^{(d)}_j)$, which naturally decomposes into shared and domain-specific effects (full derivation in  Appendix \ref{app:approximating}). Example explanations are presented in Figure~\ref{fig:mechanisms} and Appendix~\ref{app:explanations}, where lighter bars indicate the shared contribution to the lift, while darker bars and arrows indicate domain-specific contributions.
\section{Experimental Setup}
\label{sec:experimental}

In this section, we describe our six evaluation settings and provide details about the preference mechanisms we explain, the baselines used for comparison, the data, and the training and evaluation procedures. As part of our pipeline, we use Gemini-1.5-Pro \citep{gemini-1.5} for concept discovery and representation. Full implementation details, including hyperparameters, prompts, and additional setup information are provided in Appendix \ref{sec:implementation}.

\subsection{Evaluations}
\label{sub:evaluations}

We evaluate both the method and the resulting explanations across six evaluation settings. The first three assess our modeling choices, highlighting the methodological contributions.
The latter three settings focus on evaluating explanations, a challenging problem in NLP \citep{JacoviG20, MadsenRC23}. Since explanation quality is difficult to quantify directly, we take an application-driven approach. If explanations can identify concepts that improve downstream performance compared to less important concepts, this provides indirect evidence of their faithfulness and utility.

\paragraph{Human Evaluation of Concept Representations} As part of our method, concepts are annotated by an LLM. We recruited six human annotators to validate this step and asked each to annotate 400 concepts. We measure alignment between the annotators and the LLM, and statistically validate using LLM annotations over human ones.

\paragraph{Prediction Strength} Our explanations rely on a white-box model trained to imitate preference decisions. If the model performs poorly, the explanations may be considered unreliable. We hence evaluate the accuracy of our model and compare it to other alternatives, including multiple state-of-the-art LLM-as-judges and reward models, and LLMs fine-tuned on the same data. We show that our explainable model outperforms all alternatives.

\paragraph{Ablation Study} We perform an ablation study of the HMDR model in both in-domain and out-of-domain settings. We compare our model to variants, demonstrating that our full objective yields robust performance across twelve mechanisms.

\paragraph{Hacking Judges with Explanations} To test whether our explanations identify concepts that truly matter to LLM judges, we use them to guide response generation. Specifically, we prompt LLMs to produce responses conditioned on the top-ranked concepts from a given judge. We find that LLM-as-a-Judge models consistently prefer these explanation-guided responses over other responses.

\paragraph{Improving Judges with Explanations} 
We apply our explanations to resolve tie cases where LLM-as-a-Judge models give inconsistent predictions depending on response order. We use explanations to identify the most important concepts to humans and re-prompt the judge to resolve ties based on these concepts. This procedure leads to consistent improvements in alignment with human preferences.

\paragraph{Analyzing the Explanations} Finally, we analyze global explanations. We find that the effects of our automatically discovered concepts align with and extend prior studies of manually curated concepts. This supports our method's validity and ability to recover and build upon existing insights.

\subsection{Models}
\label{sub:models}

The models we explore serve two purposes: (1) as mechanisms we explain; (2) as baselines for preference prediction, against which we compare our method. Notably, our method is the first complete pipeline for preferences explainability, from concept discovery to modeling, so comparisons focus on prediction, where established alternatives exist, rather than explainability.


\paragraph{LLMs-as-Judges} We evaluate six LLMs: GPT-4o and GPT-4o-mini \citep{gpt-4o}, Gemini-1.5-Pro and Gemini-1.5-Flash \citep{gemini-1.5}, and Llama-3.1-8B-Instruct \citep{llama3.1}. In addition to zero-shot settings, we experiment with Chain-of-Thought (CoT) prompting for each LLM, and with few-shot prompting for Gemini-1.5-Flash only (due to high computational costs). Since LLMs can be sensitive to the order in which responses are presented, the evaluations are conducted with response positions swapped, and if the predictions differ between orders, the instance contributes 0.5 to the overall accuracy. Few-shot accuracy scores are computed using an ensemble of eight randomly sampled demonstration sets, with the final prediction determined by majority vote. 

\paragraph{Reward Models} We explore two reward models, QRM \citep{qrm} and Skywork \citep{skywork}, which at the time of writing, are the two best-performing 8B-parameter models on the RewardBench leaderboard.\footnote{\url{https://huggingface.co/spaces/allenai/reward-bench}} Finally, we also experiment with six encoder models and LLMs fine-tuned on our dataset, see Appendix \ref{sub:hyperparameters}.


\paragraph{Ablation Models} 
We compare the HMDR model to several white-box alternatives: (1) \textit{Shared Model:} a logistic regression model trained only on shared concepts, without any domain-specific deviations; (2) \textit{Specific Model:} a domain-specific logistic regression model that learns separate weights for each domain, without shared parameters; and (3) \textit{Dirty Model:} a binary classification variant of the dirty model from \citet{JalaliRSR10}, which includes shared weights and domain-specific deviations but lacks our shared loss objective. 

\subsection{Data and Training Setups}
\label{sub:data}

\paragraph{Dataset} Our dataset spans eight diverse domains, curated from various preference data sources. Each domain contains 800 examples (400 for concept discovery, and 400 for training and testing models). As shown in our results, these domains are challenging and not saturated, in contrast to related benchmarks such as \citet{Lambert2024RewardBench}. 

Five domains are sourced from Reddit \citep{shp}: \textit{General, Travel, Food, Legal,} and \textit{Picks} (book and movie recommendations). Each domain includes posts (user queries) from subreddits focused on topics related to the domain name. The preference labels are derived from community upvotes: the chosen response must have at least 15 upvotes, at least twice as many as the rejected response, appear later in the thread, and be of similar length to the rejected one.

The sixth domain, \textit{Software}, is based on StackOverflow and focuses on software-related questions. It follows the same preference selection criteria as the Reddit domains. We also include two RLHF datasets: \textit{PKU}, a safety-focused preference dataset \citep{pku}, and \textit{UltraFeedback (UFB)}, a general RLAIF dataset  \citep{ultrafeedback}.\footnote{The preferences in UFB are based on GPT-4.}

\paragraph{In-Domain and Out-of-Domain} In-domain evaluation uses all domains for training, with results averaged over 25 random train ($n=2800$) test ($400$) splits. For out-of-domain evaluation, we adopt a leave-one-out setup, training on seven domains and testing on the held-out one, repeated across five seeds per domain. Each split/seed includes hyperparameter tuning via 5-fold cross-validation. 
\section{Results}
\label{sec:results}

\subsection{Method Evaluation}
\label{sub:method_eval}

\paragraph{Human Evaluation} To assess whether the LLM (Gemini-1.5-Pro) can reliably annotate concepts and represent triplets, we conducted a blind human evaluation study with six annotators, each labeling $400$ concepts. We applied the Alternative Annotator Test \citep{Calderon25Alternative}, a statistical procedure that evaluates whether LLMs can replace human annotators. The LLM achieves an advantage probability of $\rho = 0.85$, indicating its annotations were as good as or better (closer to majority vote) than those of humans $85\%$ of the time. It also passes the test at the conservative threshold $\varepsilon = 0.1$ (amount of acceptable disagreements), validating the usage of LLM annotations. Additional details are provided in Appendix~\ref{app:human_eval}.

\begin{figure}[t]
    \centering
    \includegraphics[width=0.465\textwidth]{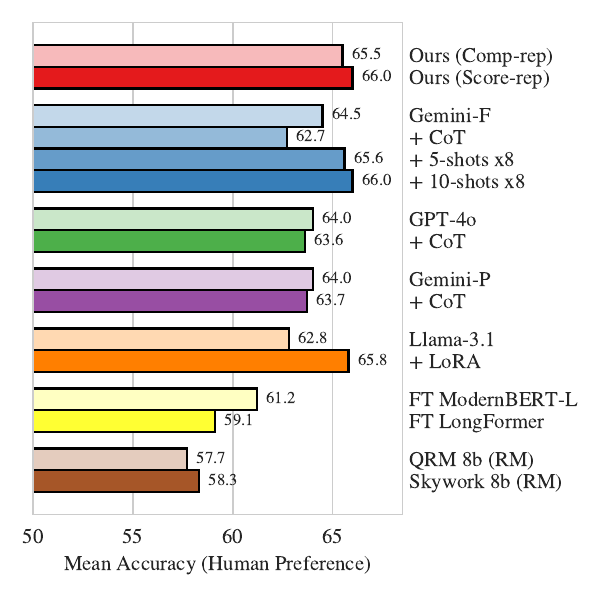}
    \vspace{-1.2em}
    \caption{\textbf{Human Preference Results:} Average accuracy across eight domains. Unlike other baselines, our method is explainable while also achieving performance comparable to the strongest LLM-as-a-Judge: an ensemble of Gemini-Flash using eight 10-shot prompts.}
    \label{fig:human_preference}
\vspace{-0.8em}
\end{figure}
\begin{table}[t]
\centering
\footnotesize
\begin{adjustbox}{width=0.495\textwidth}
\begin{tabular}{l|cc|ccc|cc}
\toprule
\multirowcell{2}{\textbf{Explained} \\ \textbf{Mech. $\downarrow$}} & \multicolumn{2}{c|}{\textbf{Ours}} & \multicolumn{2}{c}{\textbf{Shared}} & \textbf{Spe.} & \multicolumn{2}{c}{\textbf{Dirty}} \\
& \underline{In} & \underline{Out} & \underline{In} & \underline{Out} & \underline{In} & \underline{In} & \underline{Out} \\
\midrule
Human                &             \textcolor{Green}{\textbf{65.5}} &                 \textcolor{Purple}{\textcolor{Purple}{\textbf{62.9}}} &               62.3 &                   62.3 &                 63.6 &              65.1 &                  62.1 \\
\midrule
Gemini-F             &             \textcolor{Green}{\textbf{82.8}} &                 81.6 &               82.2 &                   \textcolor{Purple}{\textbf{81.8}} &                 82.6 &              \textcolor{Green}{\textbf{82.8}} &                  79.3 \\
\null\quad + 10-shots &             78.2 &                 \textcolor{Purple}{\textbf{78.1}} &               77.9 &                   77.4 &                 77.8 &              \textcolor{Green}{\textbf{79.2}} &                  76.7 \\
Gemini-P             &             84.1 &                 \textcolor{Purple}{\textbf{83.4}} &               \textcolor{Green}{\textbf{84.3}} &                   83.0 &                 84.2 &               83.4 &                  82.4 \\
\null\quad +  CoT         &             \textcolor{Green}{\textbf{85.0}} &                 \textcolor{Purple}{\textbf{82.5}} &               83.9 &                   82.3 &                 84.8 &               84.6 &                  81.1 \\
\midrule
GPT-4o-mini          &             79.6 &                 \textcolor{Purple}{\textbf{79.3}} &               \textcolor{Green}{\textbf{79.8}} &                   78.9 &                 79.1 &              79.4 &                  78.0 \\
\null\quad +  CoT      &             \textcolor{Green}{\textbf{81.3}} &                 \textcolor{Purple}{\textbf{80.1}} &               80.6 &                   79.6 &                 80.8 &              80.8 &                  79.7 \\
GPT-4o               &             \textcolor{Green}{\textbf{80.0}} &                 \textcolor{Purple}{\textbf{79.1}} &               79.5 &                   \textcolor{Purple}{\textbf{79.1}} &                 79.3 &               \textcolor{Green}{\textbf{80.0}} &                  76.5 \\
\null\quad +  CoT           &             83.2 &                 \textcolor{Purple}{\textbf{82.6}} &               82.6 &                   82.5 &                 83.5 &              \textcolor{Green}{\textbf{83.7}} &                  81.0 \\
\midrule
Llama-3.1            &             \textcolor{Green}{\textbf{81.6}} &                 \textcolor{Purple}{\textbf{79.2}} &               80.6 &                   79.0 &                 81.0 &              80.9 &                  79.0 \\
QRM 8b               &             69.6 &                 68.9 &               69.3 &                   \textcolor{Purple}{\textbf{69.3}} &                 69.5 &              \textcolor{Green}{\textbf{70.2}} &                  66.9 \\
Skywork 8b           &             \textcolor{Green}{\textbf{69.7}} &                 \textcolor{Purple}{\textbf{69.1}} &               69.4 &                   69.2 &                 69.1 &              \textcolor{Green}{\textbf{69.7}} &                  67.6 \\
\midrule
\textbf{Mean} & \textcolor{Green}{\textbf{78.4}} &                 \textcolor{Purple}{\textbf{77.2}} &               77.7 &                   77.0 &                 77.9 &              78.3 &                  75.9 \\
\bottomrule
\end{tabular}
\end{adjustbox}
\caption{\textbf{In-domain and Out-of-domain Results:} Each row corresponds to one of the 12 explained preference mechanisms. We report in-domain (\underline{In}) and out-of-domain (\underline{Out}) accuracies. The columns compare explainable regression models. Bold colored numbers indicate the highest \textcolor{Green}{\textbf{In (green)}} or \textcolor{Purple}{\textbf{Out (purple)}} accuracy score in each row. All models are based on Comp-reps, see Table~\ref{tab:rm_results} (Appendix) for Score-reps.}
\label{tab:laaj_results}
\vspace{-0.8em}
\end{table}

\paragraph{Prediction Strength} 
While our primary goal is explainability, we start by assessing the general preference prediction capability of our concept-based white-box model trained to imitate preferences. We first apply our method to human preferences, and compare the prediction accuracy of the white-box model to state-of-the-art black-box systems. Figure~\ref{fig:human_preference} presents the average accuracy across eight domains for this setup, with detailed results and additional baselines in Table~\ref{tab:appendix_human}. The strongest LaaJ baseline is an ensemble of eight 10-shot Gemini-1.5-Flash models, followed by a 5-shot ensemble and a LoRA fine-tuned Llama-3.1. 
Remarkably, \textbf{our method achieves accuracy comparable to the strongest baseline, while being inherently interpretable} ($66\%$ and $65.5\%$ for Comp-rep and Score-rep, respectively).
Finally, we find that CoT prompting degrades LaaJ performance, whereas few-shot prompting improves it, consistent with \citet{Calderon25Alternative}. In addition, we find that CoT prompting degrades LaaJ performance, whereas few-shot prompting improves it, consistent with \citet{Calderon25Alternative}.


We next assess the performance when explaining LaaJs and RMs. Tables~\ref{tab:laaj_results} (Comp-rep) and \ref{tab:rm_results} (Score-rep, Appendix) report average in-domain and out-of-domain accuracies (detailed results in Appendix Tables~\ref{tab:appendix_in_domain} and \ref{tab:appendix_ood}). Our method (leftmost columns) achieves a relatively high average accuracy around 80\% in-domain and out-of-domain (when excluding human preferences). To assess whether our method captures meaningful relationships between concepts and preferences, Figure~\ref{fig:co_acc} (Appendix) reports pairwise agreement between all examined mechanisms. Our method achieves higher agreement than any model pair. 
This suggests that our method captures model-specific behaviors beyond simple model–model correlations.

\paragraph{Ablation Study} In this paper, we propose the HMDR model for multi-domain learning. We next evaluate how this white-box model performs both in-domain and out-of-domain compared to other variants. The results are presented in Tables~\ref{tab:laaj_results} (Comp-rep) and \ref{tab:rm_results} (Score-rep, Appendix). As shown, the HMDR model achieves the highest average performance both in-domain and out-of-domain. In particular, for nearly every explained mechiansm, it performs the best in either in-domain or out-of-domain, often in both. While the only-shared variant performs similarly to our method out-of-domain, it underperforms in-domain. Conversely, the dirty model performs similarly in-domain but underperforms out-of-domain. Our results emphasize the advantages of the HMDR model: the hierarchical decomposition enables capturing both shared and domain-specific effects, while the optimization objective yields strong performance that supports model generalization.

\begin{table}[t]
\centering
\normalsize
\begin{adjustbox}{width=0.48\textwidth}
\begin{tabular}{lll|ccc|c}
\toprule
\multirow{2}{*}{\textbf{Judge}} & \multirow{2}{*}{\textbf{Generator}} & \textbf{Concepts} & \multicolumn{4}{l}{\textit{Comparison with \textcolor{Gray}{\textbf{vanillas}}}}\\
 & & \textit{\quad to follow} & Win & Tie & Lose & \textbf{WR} \\
\midrule
\multirow{4}{*}{Gemini-P} & \multirow{2}{*}{GPT-4o-m} & \textcolor{RoyalBlue}{\textbf{Explanation}} &  76.2 &  19.3 &   4.5 &           \textbf{85.9} \\
         &          & \textcolor{YellowOrange}{\textbf{Random}} &  51.5 &  27.3 &  21.2 &           65.1 \\
         \cmidrule{2-7}
         & \multirow{2}{*}{Gemini-F} & \textcolor{RoyalBlue}{\textbf{Explanation}} &  46.9 &  33.0 &  20.2 &           \textbf{63.4} \\
         &          & \textcolor{YellowOrange}{\textbf{Random}} &  27.7 &  32.8 &  39.5 &           44.1 \\
\midrule
\multirow{4}{*}{GPT-4o} & \multirow{2}{*}{GPT-4o-m} & \textcolor{RoyalBlue}{\textbf{Explanation}} &  38.8 &  59.8 &   1.5 &           \textbf{68.7} \\
         &          & \textcolor{YellowOrange}{\textbf{Random}} &  27.0 &  63.2 &   9.8 &           58.6 \\
         \cmidrule{2-7}
         & \multirow{2}{*}{Gemini-F} & \textcolor{RoyalBlue}{\textbf{Explanation}} &  20.1 &  74.6 &   5.3 &           \textbf{57.4} \\
         &          & \textcolor{YellowOrange}{\textbf{Random}} &  13.9 &  63.7 &  22.4 &           45.8 \\
\midrule
\multirow{4}{*}{QRM} & \multirow{2}{*}{GPT-4o-m} & \textcolor{RoyalBlue}{\textbf{Explanation}} &  54.7 &   0.7 &  44.5 &           \textbf{55.1} \\
         &          & \textcolor{YellowOrange}{\textbf{Random}} &  43.5 &   1.3 &  55.2 &           44.1 \\
         \cmidrule{2-7}
         & \multirow{2}{*}{Gemini-F} & \textcolor{RoyalBlue}{\textbf{Explanation}} &  49.7 &   0.8 &  49.5 &           50.1 \\
         &          & \textcolor{YellowOrange}{\textbf{Random}} &  35.8 &   0.5 &  63.7 &           36.0 \\
\bottomrule
\end{tabular}
\end{adjustbox}
\vspace{-0.35em}
\caption{\textbf{Judge Hack Results:} Evaluating responses with a LaaJ (`Judge') that were generated by GPT-4o-mini and Gemini-Flash (`Generator'). Responses are generated by following four top judge-selected concepts (\textcolor{RoyalBlue}{\textbf{Explanation}}) or four random concepts (\textcolor{YellowOrange}{\textbf{Random}}). The judge compared the concept-guided responses against \textcolor{Gray}{\textbf{vanillas}} (prompt without concepts). $\text{\textbf{WR}}=\text{Win}+\tfrac12\text{Tie}$. Bold numbers indicate that $\text{WR} > 50\%$ is statistically significant after Bonferroni correction ($\alpha<.006$).}
\label{tab:results_hack}
\vspace{-0.8em}
\end{table}
\begin{table}[t]
\centering
\footnotesize
\begin{adjustbox}{width=0.48\textwidth}
\begin{tabular}{l|ccc|cc}
\toprule
\multirow{2}{*}{\quad \textbf{Tie Subset of $\rightarrow$}} & \multicolumn{3}{c|}{\textbf{Gemini-1.5-Flash}} & \multicolumn{2}{c}{\textbf{GPT-4o-m}} \\
& \underline{zero} & \underline{CoT}   & \underline{10-shot}  & \underline{zero}  & \underline{CoT} \\
\midrule
\% Ties & 21.3 & 32.1 &10.7 & 19.1 & 21.5 \\
\midrule
\multicolumn{6}{c}{\textit{Accuracy gains (agreement with humans) on subsets of tie cases:}}\\
Gemini-1.5-Flash  & 0.0 & 6.2 & 2.7 & 4.7 & 4.8 \\
 + \textcolor{YellowOrange}{\textbf{Random Concepts}} & 1.8 & 3.3 & 2.1 & 4.7 & 4.0 \\
 + \textcolor{RoyalBlue}{\textbf{Gemini-F Exp.}}  & 2.2 & 5.5 & 3.3 & 5.4 & 5.2 \\
 + \textcolor{OliveGreen}{\textbf{Human Exp.}}  & 2.9 & 5.7 & 5.2 & 5.3 & 6.3 \\
 + \textcolor{WildStrawberry}{\textbf{Diff Exp.}}  & \textbf{5.2} & \textbf{7.8} & \textbf{4.0} & \textbf{6.3} & \textbf{7.8} \\
\midrule
GPT-4o-mini  & 1.7 & 4.9 & 4.2 & 0.0 & 1.5 \\
 + \textcolor{YellowOrange}{\textbf{Random Concepts}} & 1.7 & 3.4 & 5.2 & 3.5 & 2.1 \\
 + \textcolor{RoyalBlue}{\textbf{GPT-4o-m Exp.}}  & 2.9 & 5.1 & 7.3 & 2.2 & 2.1 \\
 + \textcolor{OliveGreen}{\textbf{Human Exp.}}  & 4.7 & 5.4 & 5.4 & 5.4 & 3.8 \\
 + \textcolor{WildStrawberry}{\textbf{Diff Exp.}}  & \textbf{6.6} & \textbf{7.3} & \textbf{10.8} & \textbf{6.1} & \textbf{4.7} \\
\bottomrule
\end{tabular}
\end{adjustbox}
\caption{\textbf{Tie-Breaking Gain:} Columns represent the examined tie subsets of the specified LaaJ. Rows report accuracy \textbf{gains ($\Delta$\%)} when resolving ties using Gemini-1.5-Flash, GPT-4o-mini, and different strategies. Ties are resolved based on which response better follows four \textcolor{YellowOrange}{\textbf{random}} concepts, or concepts with the largest weights according to the \textcolor{RoyalBlue}{\textbf{judge's explanation}}, \textcolor{OliveGreen}{\textbf{human explanation}}, or the \textcolor{WildStrawberry}{\textbf{difference}} between them.}
\label{tab:results_ties}
\vspace{-0.8em}
\end{table}

\subsection{Explanations Evaluation}
\label{sub:exp_eval}

\paragraph{Hacking Judges} Our first application-driven evaluation is \textit{Judge Hack}, which leverages explanations of an LaaJ or an RM to guide another LLM in generating responses. If the judge truly relies on concepts identified as important by our explanations, then prompting the generator to align with those concepts should improve its ranking. 

To test this, we use two LaaJs (Gemini-1.5-Pro and GPT-4o) and a reward model (QRM), along with two generator LLMs (Gemini-1.5-Flash and GPT-4o-mini). We sample from each domain 50 queries (400 total) not seen during explanation training, and generate a \textcolor{Gray}{\textbf{vanilla response}} using the generators. We then extract the top four concepts (largest weights) per domain from judge explanations and generate \textcolor{RoyalBlue}{\textbf{explanation responses}} by prompting the generator to \textit{``consider the following concepts when responding.''} As a control, we also generate \textcolor{YellowOrange}{\textbf{random responses}} using four randomly selected domain-specific concepts. Each response (explanation or random) is compared to the \textcolor{Gray}{\textbf{vanilla responses}}, resulting in 4,800 comparisons.

As shown in Table~\ref{tab:results_hack}, \textcolor{RoyalBlue}{\textbf{explanation-guided responses}} are consistently preferred over \textcolor{Gray}{\textbf{vanilla responses}}, with a win rate $\text{WR}=\text{Win}+\tfrac12 \text{Tie} > 50\%$, and by a much larger margin than \textcolor{YellowOrange}{\textbf{random responses}}. For Gemini-1.5-Pro as judge, the win rate improves by +20.8 and +19.3 points over \textcolor{YellowOrange}{\textbf{random responses}}, and for GPT-4o, by +10.1 and +11.6 points. These improvements indicate that our explanations capture relevant concepts that meaningfully influence judge behavior.

\paragraph{Breaking Ties} In the second application-driven evaluation, \textit{Tie Break}, we use explanations to resolve cases where the LaaJ prediction flips depending on the order of responses in the prompt.
Rather than asking the LaaJ which response is better, we re-prompt it to decide which response better aligns with concepts important to humans. If this improves alignment with them, it suggests our explanations capture meaningful concepts.

\begin{figure*}[t]
    \centering
    \includegraphics[width=0.95\textwidth]{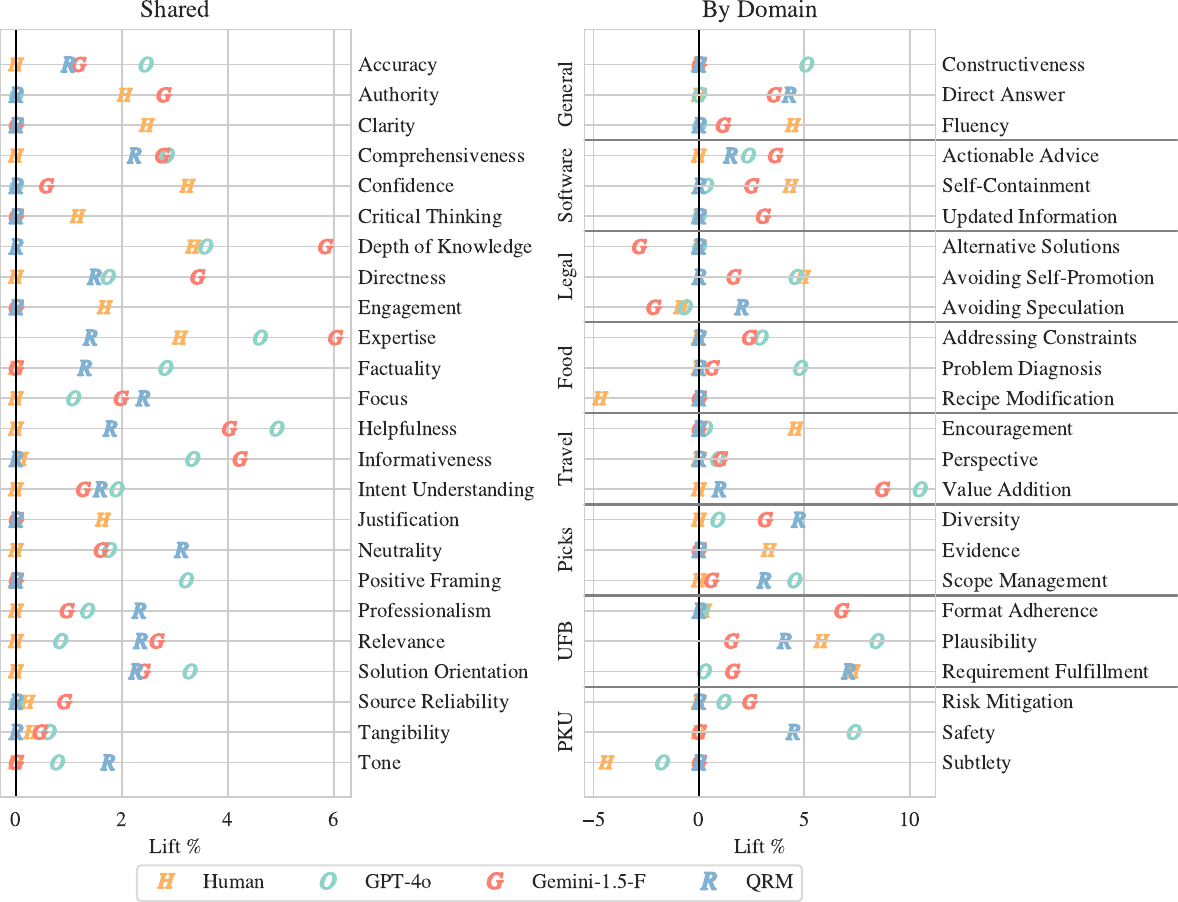}
    \caption{\textbf{Explanations Analysis:} Lifts of four mechanisms: human preferences (yellow $H$), GPT-4o (green $O$), Gemini-1.5-Flash (red $G$), and QRM (blue $R$). Right: shared contributions of 24 concepts, selected from the top ten shared weights of the four mechanisms. Left: For each domain, we select three concepts from a set of four not shared concepts that have the highest domain-specific weight in at least one of the four explanations.}
    \label{fig:exp_analysis}
\end{figure*}

To examine the Tie Break procedure, we employ two LaaJs, Gemini-Flash and GPT-4o-mini, across three configurations (zero-shot, few-shot, and CoT). We train two explanation models using only non-tie examples (one for the LaaJ and one for humans). We then extract the four top concepts (largest weights) from: the \textcolor{RoyalBlue}{\textbf{LaaJ explanation}}, the \textcolor{OliveGreen}{\textbf{human explanation}}, and the \textcolor{WildStrawberry}{\textbf{differences}} between them (i.e., human weight minus judge weight). The idea behind the latter approach is that weight differences highlight which concepts the judge should focus on when it currently does not. As a baseline, we also consider tie-breaking using \textcolor{YellowOrange}{\textbf{randomly}} sampled concepts.  We focus on the subset of tie cases for each LaaJ and use concept-guided prompts to resolve them, reporting accuracy gains (agreements with humans) over not resolving ties.

Table~\ref{tab:results_ties} presents accuracy gains on the tie subsets. We observe a consistent and meaningful trend across all judges. The accuracy ranking follows: random $\le$ LaaJ $\le$ human $\le$ differences, which aligns with expectations. \textcolor{YellowOrange}{\textbf{Random concepts}} have the weakest effect, while \textcolor{RoyalBlue}{\textbf{judge-explanation concepts}} reinforce what the judge already considers. \textcolor{OliveGreen}{\textbf{Human-explanation concepts}} emphasize factors that align with human preferences, but the \textcolor{WildStrawberry}{\textbf{difference-based concepts}} prompt the judge to focus on overlooked human-aligned factors.

\paragraph{Analysis of Explanations} 
The goal of this analysis is to evaluate our explanations by comparing our automatically discovered concepts to manually curated ones from prior studies. We begin by assessing how well the effects of our shared concepts recover past findings. We then examine the added value of scalable, domain-specific concept discovery by analyzing the frequency and prominence of these concepts in our explanations.

Figure~\ref{fig:exp_analysis} illustrates the impact of the 24 most influential shared concepts (right) and 24 domain-specific concepts (left) for four mechanisms: humans, GPT-4o, Gemini-1.5-Flash, and QRM reward model. 
We begin by examining shared concepts and observe that different preference mechanisms prioritize different aspects, with some concepts highly weighted by one mechanism but irrelevant to others. This aligns with findings from \citet{li2024dissecting}, who analyzed manually curated, domain-agnostic concepts. For human preferences, we find a strong emphasis on `Clarity', `Authority', and `Confidence', consistent with prior work identifying `clear', `well-formatted' \citep{li2024dissecting}, and `Authoritative' as key concepts \citep{sharma2023towards}.
Both humans and LLMs assign high importance to `Depth of Knowledge' and `Expertise', however, `Accuracy' and `Factuality' is among the top concepts for GPT-4o but receives no weight from human annotators (`no severe error' is the leading concept for GPT-4 in \citet{li2024dissecting}). One interpretation is that non-expert human evaluators (such as those in our Reddit-based dataset) favor responses that appear knowledgeable and expert-like, but are less able to verify their content, unlike LLMs, which are better in this task \citep{nahum2024llms}.
Finally, `Helpfulness' ranks highly for LLMs, likely reflecting the objectives of alignment research \citep{Bai2022Helpfulness}.

We next discuss the domain-specific concepts. Our results reveal that many of the most important concepts influencing preferences 
are domain-specific. This is very observable in the explanations in Appendix \ref{app:explanations}, where domain-specific concepts (concepts with only dark bars and arrows) are at the top of most domains and mechanisms. The following dual observation highlights both our method's necessity and effectiveness: it captures domain-specific nuances while maintaining consistency with broadly accepted quality criteria in the literature, offering a scalable and generalizable approach to modeling preferences in NLP.

\section{Conclusions}
\label{sec:conclusions}

In this work, we explored a new paradigm for concept-based explainability of preferences, involving automatic concept discovery, concept-based representations of examples, and multi-domain modeling using a white-box HMDR model.
We demonstrated how this approach can be evaluated, including application-driven evaluations. 
We hope this work will inspire others and support the scalable explainability of core components in LLM research and development, such as preferences.
\section{Limitations}
\label{sec:limitations}



\paragraph{Linear Model} The HMDR is a linear model, in which the relationship between concepts and preferences is modeled using a linear function, which may not fully capture the complex, nonlinear decision processes underlying human or LLM preference mechanisms. However, this linearity is not merely a limitation but also a deliberate design choice: linear white-box models are far more interpretable to humans, especially when constrained to a few features. Prior work has shown that such models improve human understanding \citep{poursabzi2021manipulating}. 

\paragraph{Causality and Faithfulness} Faithful explanations of mechanisms requires causality \citep{JacoviG20, GatCFCSR24}. However, our method is not causal: it does not identify or account for the underlying causal structure governing the relationship between concepts and preferences. That said, logistic regression can still offer a useful approximation under standard assumptions, particularly when relevant confounders are included \citep{cinelli2024crash}. While this does not substitute for formal causal inference frameworks such as randomized controlled trials, it provides a starting point. A promising direction for future work is to discover the causal graph over the concepts that drive preference mechanisms.

\paragraph{High Computational Costs} Another limitation of our method is its reliance on many LLM calls, from concept discovery to concept-based representations of triplets. When global explanations are of interest, the concept discovery and representation steps need only be performed once. The resulting representations can then be reused to train white-box models under different preference labels, enabling efficient explanation of a broad range of mechanisms at the cost of a single discovery and representation phase. However, the computational overhead is particularly pronounced in the case of ``real-time'' local (per-example) explanations, which require representing the new triplet. Yet, the HMDR model encourages sparsity and in practice, only a small subset of concepts have non-zero weights (see Figure~\ref{fig:hparams_analysis}), which helps reduce costs. 

Nevertheless, for preference prediction (although our focus is not on prediction but on explainability), our method requires more computational effort than standard zero-shot prompting. It involves generating concept-based representations using prompts that include concept definitions, resulting in longer inputs. For this reason, we also evaluate a few-shot ensemble model (eight calls with 10 shots each), which serves as a more computationally comparable baseline due to its reliance on multiple inferences with long inputs.

\paragraph{Explainability Evaluation} Evaluating explanations in NLP remains a fundamentally challenging and open problem, particularly for concept-level explainability, which is less explored than token-level approaches. The lack of standardized, widely accepted evaluation metrics limits our ability to make definitive claims about explanation quality. In this work, however, we assess the usefulness of our explanations through two novel, application-driven evaluation settings: Judge Hack and Tie Break. Our results show that the identified concepts improve downstream performance compared to less important ones, providing indirect evidence of explanation quality.

\section*{Acknowledgments}
This research is supported by the IBM–Technion Research Collaboration.

\bibliography{custom, anthology}

\clearpage

\appendix

\renewcommand \thepart{}
\renewcommand \partname{}
\mtcsettitle{parttoc}{}
\addcontentsline{toc}{section}{Appendix} 
\part{Appendix} 
\parttoc 


\section{Background}
\label{app:background}

\subsection{Concept-based Explainability}
\label{app:back_exp}

While the terms interpretability and explainability are often used interchangeably in NLP research \citep{miller2019explanation, JacoviG20, lyu2024faithfulness, calderon2025stakeholders}, the works of \citet{lipton2016mythos} and \citet{doshi2017towards} emphasize the importance of definitional clarity, distinguishing between \textit{interpretability}, how understandable a model is to humans, and \textit{explainability}, which refers to post-hoc explanations of model predictions. In this work, we focus on the explainability of preference mechanisms using an interpretable white-box model. Specifically, we emphasize human-interpretable, concept-based explanations. Concept-based explanations support communicating insights in understandable terms for any stakeholder \citep{calderon2025stakeholders}. 
Unlike token-level methods such as feature attributions or attention-based explanations \citep{LuoIHP24, ZhaoCYLDCWYD24, calderon2025stakeholders}, concept-based explanations more closely resemble human reasoning \citep{kim2018tcav, alqaraawi2020cnnuserstudy, KimWRFM23, poeta2023conceptxai}, facilitate abstraction \citep{FederOSR21}, reduce the cognitive load of explaining lengthy raw textual inputs \citep{calderon2025stakeholders}, and naturally support both \textit{local explanations}, describing the mechanism for an individual example, and \textit{global explanations}, describing the mechanism over a distribution of examples \citep{GatCFCSR24}. One approach to concept-based explainability is concept bottleneck models \citep{koh2020conceptbottleneck}, which use interpretable concepts as intermediate variables. Like other recent work \citep{ludan2023tbm, sun2024cblm}, our method leverages LLMs to discover such concepts. However, while prior studies focus on traditional tasks such as sentiment analysis or topic classification, our work targets general \textit{preference mechanisms} in a multi-domain learning setting.

\subsection{Multi-domain Learning}
\label{app:back_md}

Multi-domain learning aims to train models that perform well across multiple domains, where both input and output distributions may shift, and potentially generalize better to unseen domains \citep{daume2007frustratingly, ben2010theory, Ben-DavidOR22, CalderonBFR22}.
One common approach to multi-domain learning focuses on learning domain-invariant representations, emphasizing shared features to improve domain robustness \citep{CalderonPBCGOSR24}. This includes methods such as pivot features \citep{blitzer2006domain, ziser2018pivot, Ben-DavidRR20}, domain adversarial networks (DANN) \citep{ganin2016domain, LiBC18}, and invariant risk minimization (IRM) \citep{arjovsky2019invariant, PeyrardGJAPCKT022}.
In contrast, another approach focuses on learning domain-specific representations, allowing models to specialize their predictions for each domain. For example, mixture-of-experts (MoE) \citep{gururangan2021demix, Ben-DavidOR22, Chirkova2024Invest} and, at the extreme, hypernetworks, which generate domain-specific weights even for unseen domains \citep{VolkBACR23, ChauhanZLMC24}.
Our HMDR model combines both approaches and is inspired by extensive work in multi-task learning, where models learn shared representations across tasks while allowing for task-specific specialization 
\citep{CollobertW08, Ruder17a, RotmanR22, ChenZY24}. However, our multi-domain learning model addresses a single task, requiring the shared weights themselves to be predictive rather than merely supportive of task-specialized components. We are also inspired by the Dirty Model from multi-task learning for regression tasks \citep{JalaliRSR10}, which decomposes model weights into shared and task-specific components. Our HMDR model, designed for classification tasks, is optimized for domain generalization and supports different sparsity structures.

\section{Additional Results}

\subsection{Human Evaluation}
\label{app:human_eval}

To assess whether the LLM (Gemini-1.5-Pro) can reliably annotate concepts and represent triplets, we conducted a blind human evaluation study. Six human annotators independently labeled a subset of the dataset (40 triplets $\times$ 10 concepts, totaling $N=400$ annotations per annotator). Each annotator was presented with a triplet (with randomized response order) and ten concepts. For each concept, annotators were asked to determine whether the first response better aligns with its definition, the second response does, both align equally or the concept is not relevant to the triplet.
The annotators included two PhD students and four fourth-year undergraduate students (one female and five males, aged 24–35). Human annotators were offered course credit. Inter-annotator agreement, measured using Cohen’s $\kappa$, yielded an average pairwise score of $\kappa = 0.27$, indicating fair agreement, particularly given the subjectivity of preference annotation tasks \citep{RottgerVHP22, LissakCSOFKR24}. For comparison, the agreement between the LLM and the human majority vote was $\kappa = 0.33$.

We then apply the Alternative Annotator Test (alt-test) of \citet{Calderon25Alternative}, a statistical procedure designed to assess whether LLMs can reliably substitute for human annotators. The LLM achieves an advantage probability of $\rho = 0.85$, meaning that 85\% of its annotations are as good as or better (i.e., closer to the human majority vote) than those of individual annotators. Importantly, the LLM also passes the alt-test at $\varepsilon = 0.1$, a cost-benefit hyperparameter that controls the acceptable level of disagreement between the LLM and humans. As noted by \citet{Calderon25Alternative}, $\varepsilon = 0.1$ is a conservative setting, thus, passing the test under this threshold provides \textit{a strong statistical justification} for using LLM annotations.

\begin{table}[t]
\centering
\footnotesize
\begin{adjustbox}{width=0.495\textwidth}
\begin{tabular}{l|cc|ccc|cc}
\toprule
\multirowcell{2}{Explained \\ Mech} & \multicolumn{2}{c|}{\textbf{Ours}} & \multicolumn{2}{c}{\textbf{Shared}} & \textbf{Spe} & \multicolumn{2}{c}{\textbf{Dirty}} \\
& \underline{In} & \underline{Out} & \underline{In} & \underline{Out} & \underline{In} & \underline{In} & \underline{Out} \\
\midrule
Human                &           \textcolor{Green}{\textbf{66.0}} &               \textcolor{Purple}{\textbf{63.3}} &             63.6 &                 62.8 &               64.1 &            65.0 &                63.2 \\
\midrule
Gemini-F             &           81.1 &               81.5 &             81.1 &                 81.2 &               80.1 &            \textcolor{Green}{\textbf{81.9}} &                \textcolor{Purple}{\textbf{81.7}} \\
\null\quad + 10-shots x8 &           \textcolor{Green}{\textbf{77.8}} &               \textcolor{Purple}{\textbf{77.9}} &             77.4 &                 77.5 &               76.5 &           \textcolor{Green}{\textbf{77.8}} &                \textcolor{Purple}{\textbf{77.9}} \\
Gemini-P             &           82.2 &               \textcolor{Purple}{\textbf{82.3}} &             \textcolor{Green}{\textbf{83.0}} &                 82.0 &               82.8 &             82.5 &                81.6 \\
\null\quad + CoT         &           82.6 &               \textcolor{Purple}{\textbf{82.1}} &             82.3 &                 81.8 &               \textcolor{Green}{\textbf{82.7}} &             82.3 &                81.4 \\
\midrule
GPT-4o-mini          &           78.4 &               \textcolor{Purple}{\textbf{78.0}} &             78.6 &                 77.9 &               78.1 &            \textcolor{Green}{\textbf{78.7}} &                77.4 \\
\null\quad + CoT      &           78.4 &               \textcolor{Purple}{\textbf{78.9}} &             \textcolor{Green}{\textbf{79.1}} &                 78.7 &               79.0 &            78.4 &                77.2 \\
GPT-4o               &           \textcolor{Green}{\textbf{79.9}} &               79.4 &             \textcolor{Green}{\textbf{79.9}} &                 78.9 &               79.3 &             79.4 &                \textcolor{Purple}{\textbf{79.6}} \\
\null\quad + CoT           &           81.6 &               \textcolor{Purple}{\textbf{81.7}} &             81.6 &                 \textcolor{Purple}{\textbf{81.7}} &               80.6 &            \textcolor{Green}{\textbf{82.0}} &                80.7 \\
\midrule
Llama-3.1            &           \textcolor{Green}{\textbf{78.7}} &               \textcolor{Purple}{\textbf{78.8}} &             78.6 &                 78.4 &               78.3 &            78.5 &                78.5 \\
QRM 8b               &           \textcolor{Green}{\textbf{68.9}} &               \textcolor{Purple}{\textbf{69.4}} &             \textcolor{Green}{\textbf{68.9}} &                 69.2 &               68.5 &            \textcolor{Green}{\textbf{68.9}} &                69.2 \\
Skywork 8b           &           68.3 &               68.2 &             \textcolor{Green}{\textbf{68.9}} &                 \textcolor{Purple}{\textbf{68.6}} &               68.7 &            67.7 &                68.0 \\
\midrule
\textbf{Mean}                 &           \textcolor{Green}{\textbf{77.0}} &               \textcolor{Purple}{\textbf{76.8}} &             76.9 &                 76.6 &               76.6 &            76.9 &                76.4 \\
\bottomrule
\end{tabular}
\end{adjustbox}
\caption{\textbf{In-domain and Out-of-domain Results (Score-rep):} Each row corresponds to one of the 12 explained preference mechanisms. We report in-domain (\underline{In}) and out-of-domain (\underline{Out}) accuracies. The columns compare explainable regression models. Bold colored numbers indicate the highest \textcolor{Green}{\textbf{In (green)}} or \textcolor{Purple}{\textbf{Out (purple)}} accuracy score in each row. All models are based on Score-reps, see Table~\ref{tab:laaj_results} for Comp-reps.}
\label{tab:rm_results}
\vspace{-0.8em}
\end{table}
\begin{figure}[t]
    \centering
    \includegraphics[width=0.475\textwidth]{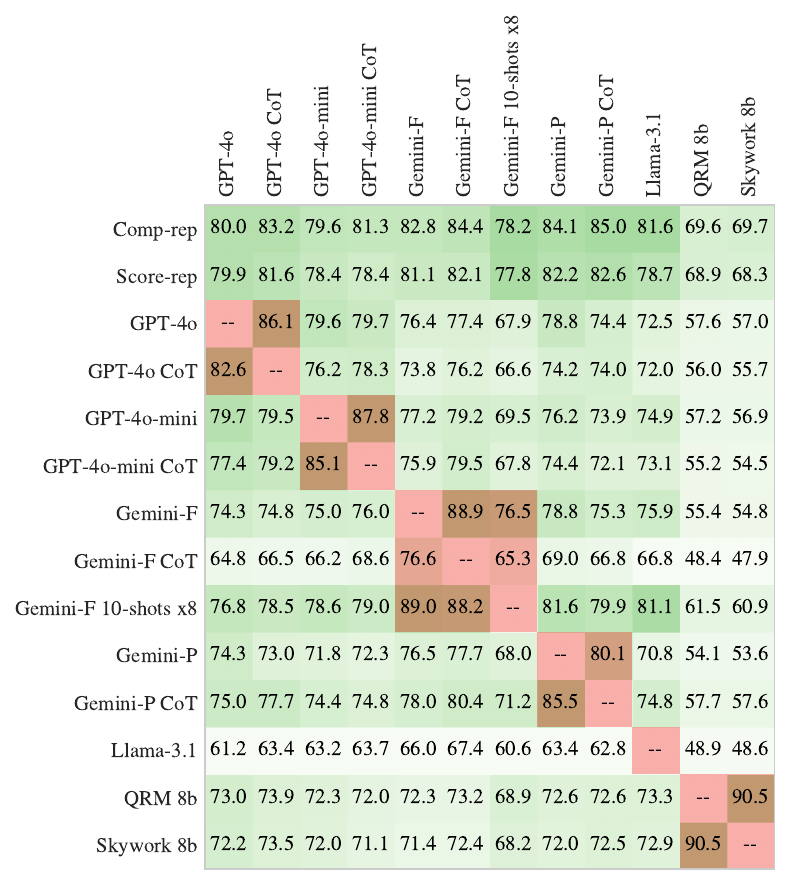}
    \caption{\textbf{Model Agreements:} Columns represent the gold labels (with tie cases removed). Each row shows the accuracy of a model against these labels. Our method (top two rows) achieves higher accuracy than other models, excluding those from the same family (highlighted in red). This demonstrates that the predictive power of our method stems not only from model-model correlations but also from capturing model-specific nuances.}
    \label{fig:co_acc}
\vspace{-0.8em}
\end{figure}

\subsection{Prediction Performance}
\label{app:prediction}

In this subsection, we provide additional tables and figures presenting results from our preference prediction performance experiments:
\begin{itemize}
\item Table~\ref{tab:rm_results}: Summary of Score-rep used with the HMDR model, and ablation models when explaining twelve mechanisms, both in-domain and out-of-domain.
\item Figure~\ref{fig:co_acc}: Agreement matrix between all pairs of mechanisms, including our method. Rows indicate the agreement accuracy when one model predicts the preferences of another.
\item Table~\ref{tab:appendix_human}: Complete results for predicting human preferences, broken down by domain. Includes all methods and baselines, including those not reported in the main text.
\item Table~\ref{tab:appendix_in_domain}: Full in-domain results of our method for explaining twelve mechanisms, with a domain-level breakdown.
\item Table~\ref{tab:appendix_ood}: Full out-of-domain results of our method for explaining twelve mechanisms, with a domain-level breakdown.
\item Figure~\ref{fig:hparams_analysis}: Hyperparameter analysis showing performance and the number of non-zero weights across different parameter values.
\end{itemize}

\subsection{Computational Costs}
\label{app:computational}

Our method relies on many LLM calls, ranging from the concept discovery stage to concept-based representations of the triplets. Noteworthy, for individual preference prediction, our method requires more inference time computation than standard zero-shot prompting, as it involves predicting concept-based features, with definitions provided in the prompt, also leading to longer inputs. Notice, however, that the HMDR model is sparse, and not all concepts are used in practice (see Figure~\ref{fig:hparams_analysis} for an analysis of the number of non-zero weights). The few-shot ensemble model serves as a more computationally comparable baseline, as it also involves multiple inferences with long inputs.

The total end-to-end cost of this project is approximately 2,000 USD.

\section{Implementation Details}
\label{sec:implementation}

\subsection{Concept Discovery and Representation}
\label{app:imp_concept}

In this subsection, we provide additional implementation details for the concept discovery and representation stages of our method. In addition, we describe various attempts made during the development of our method, focusing specifically on the concept discovery stage, and motivate our choices. Throughout all experiments, we use Gemini-1.5-Pro for both stages. 

\subsubsection{Batching User Queries}
\label{app:imp_batching}

We assume that the domain of each triplet is known in advance (e.g., based on the source of the query). At the beginning of the concept discovery stage, we annotate each user query with subdomains (e.g., healthcare, technology, and Python) and tasks (e.g., question answering, explanation, summarization, and advice), separately for each domain. These subdomains and tasks are then used to batch triplets together for concept discovery. Batching triplets by subdomain or task encourages the LLM to identify low-resolution domain-specific or task-specific concepts. To find the relevant subdomains and tasks of each domain, we begin by randomly sampling 10\% of the queries. We then prompt the LLM to generate a list of relevant subdomains and tasks conditioning on a batch of $n_b = 5$ queries given in its input. The prompt used is shown in Box~\ref{box:prop_subdomains}.

\begin{figure*}[t]
    \centering
    \includegraphics[width=0.975\textwidth]{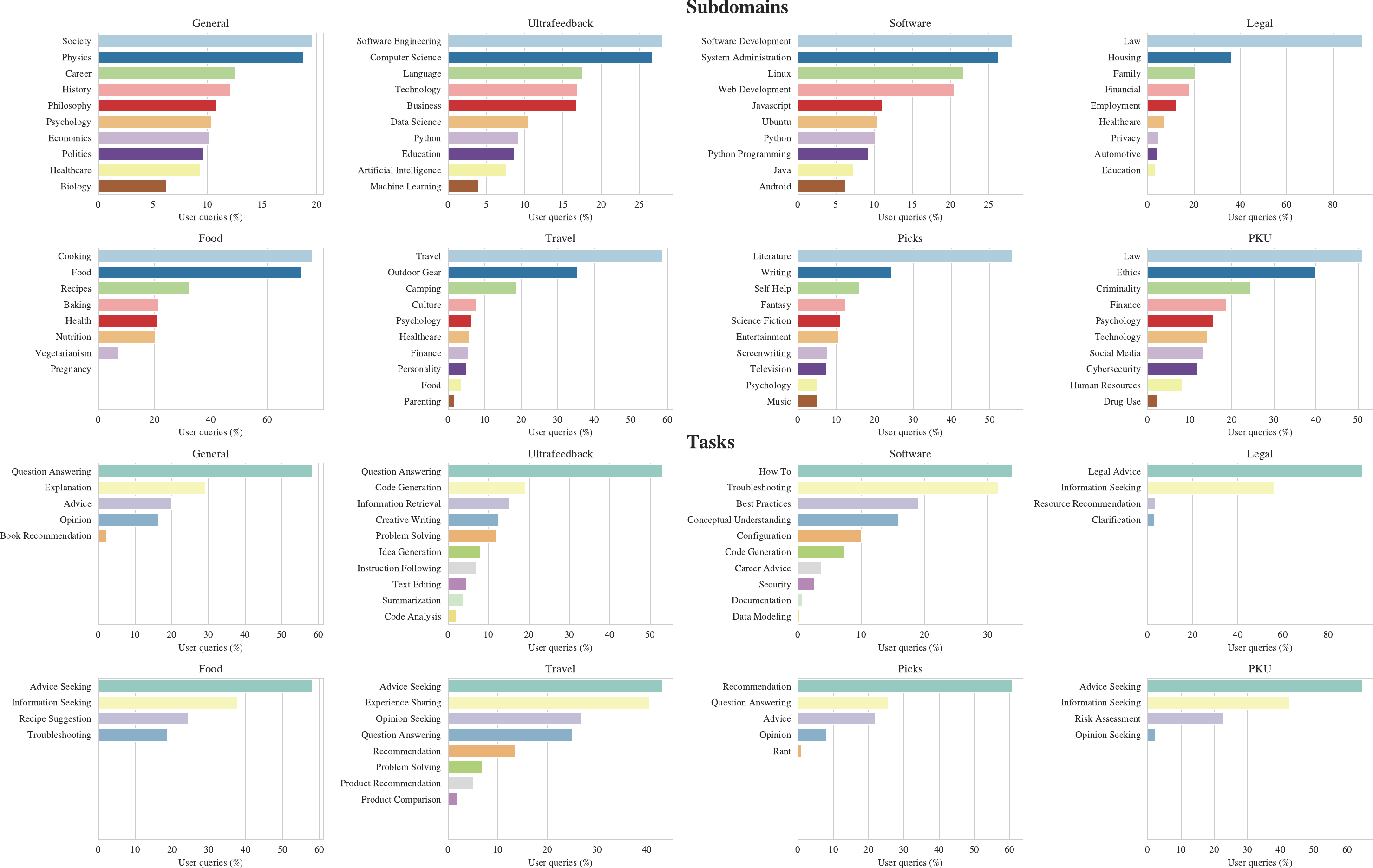}
    \caption{\textbf{Subdomains and Tasks:} For each domain, we show the percentage of user queries annotated by the LLM with each subdomain (top two rows) and task (bottom two rows). Subdomains and tasks were discovered by the LLM and used to batch examples for concept discovery. We retained up to the ten most frequent subdomains and tasks, based on annotations from 10\% of the discovery set in each domain.} 
    \label{fig:subdomains}
\vspace{-0.8em}
\end{figure*}

Next, we retain up to the ten most frequent subdomains and tasks within each domain. We then use the LLM to annotate every query $q \in d$ with subdomain(s) and task(s) from this list. The prompt used for annotation is shown in Box~\ref{box:prop_subdomains}. Figure~\ref{fig:subdomains} lists the ten subdomains and tasks for each domain, along with the proportion of queries annotated with each. Finally, we construct $B = 300$ batches of size $n_b = 5$ by randomly sampling examples that share either the same subdomain, the same task, both, or neither. Specifically, we annotate each query with a special \textit{`None'} subdomain and \textit{`None'} task. To construct a batch, we first sample a (subdomain, task) pair based on its frequency within the domain. We then randomly select $n_b = 5$ triplets, all annotated with that pair. For example, the pair (`None', `advice') refers to of all examples labeled with the `advice' task, regardless of subdomain.

\subsubsection{Discovering Concepts} 
\label{app:imp_discovery}

Each batch of triplets, along with its subdomain and task (if they are not \textit{`None'}s), is provided to the LLM for concept discovery. The LLM is tasked to propose $n_c=10$ concepts that may explain why one response is preferred over the other. Additionally, we ask the LLM to generate a concise one-sentence description of each proposed concept. These descriptions will be used later to define the concepts. To introduce variability, we slightly adjust the prompt randomly for each batch, for example, asking why the first response was chosen over the second, why the second was less favorable, or why a given response was either selected or rejected when presented alone. The prompt for concept discovery is shown in Box~\ref{box:discovery}.

During manual prompt engineering on a small subset of examples, we observed that the LLM often proposed the same set of general, non-domain-specific concepts across batches (e.g., `Helpfulness'). To address this, we manually extracted ten such common and frequent concepts and designated them as \textit{fixed concepts} (listed in Box~\ref{box:fixed_concepts}). To promote greater diversity in the concept discovery process, we modified the discovery prompt in 50\% of the batches (Box~\ref{box:discovery}), instructing the LLM to propose concepts that differ from the fixed set.

Finally, in our main setup, the preferred response for each triplet is determined by human preferences. To explore fully human-unsupervised concept discovery, we also conducted preliminary experiments using preferences from Gemini-1.5-Flash instead. We found that the vast majority of discovered concepts overlapped with those derived from human preferences. We hypothesize that this high overlap results from two factors: (1) many of the LLM and human preference labels coincide, and (2) the LLM may not rely heavily on the preference label when proposing concepts. Instead, it likely focuses on the content of the user query and responses, regardless of their ordering or preferred response. Due to budget constraints (the costliest part is concept representations), we did not complete the ablation study on the concept discovery stage. 

\subsubsection{Filtering and Defining Concepts} 
\label{app:imp_filtering}

Many of the discovered concepts are semantic duplicates. To address this, we first apply word stemming using the Snowball stemmer\footnote{\url{https://www.nltk.org/api/nltk.stem.SnowballStemmer.html}} and flag concept pairs that share at least one stemmed word as potential duplicates. We then use an LLM to make the final decision on whether each pair is indeed a semantic duplicate. The prompt we use is shown in Box~\ref{box:filter}. After filtering, we are left with $c=624$ concepts. Among them, 75 are \textit{shared concepts} that were discovered in at least half of the domains (i.e., 4). The average number of domain-specific concepts per domain is 92, with the following distribution: General=142, Legal=58, Software=65, Food=65, UFB=151, PKU=77, Travel=124, Picks=72.

For each concept, we collect up to five descriptions generated during the concept discovery phase and prompt the LLM to formulate a definition, as shown in Box~\ref{box:defining}. We define five concepts per LLM call, as we found this yields better definitions than prompting one concept at a time.

\subsubsection{Representing Triplets} 
\label{app:imp_representing}

We first predict the relevant concepts for each user query $q \in d$ to reduce the number of LLM calls for concept representation. We prompt the LLM with a list of the shared concepts and those specific to domain $d$. The prompt is shown in Box~\ref{box:relevant}. 

To represent triplets with the Comp-rep, we use the prompt in Box~\ref{box:comp_rep}, presenting both responses to the LLM. Each call includes up to 20 relevant concepts, as we found that longer prompts with too many concepts lead to more annotation errors. Additionally, since LLMs are sensitive to the order of the responses, we repeat the process with the response positions swapped (i.e., the first becomes second and vice versa). We then extract the concept annotations and assign a value of 0 if the annotations are inconsistent.

To represent triplets with the Score-rep, we use the prompt in Box~\ref{box:score_rep}, presenting one response at a time alongside the user query. We then extract the concept scores for each response and compute their difference.

\subsection{Models}
\label{sub:hyperparameters}

We evaluate our method in both in-domain and out-of-domain settings. In the in-domain setup, models are trained on examples from all eight domains and evaluated on a held-out test set containing examples from each domain. Reported results are averaged over 25 random seeds, each corresponding to a different train-test split with 2,800 training examples and 400 test examples.

\begin{figure*}[t]
    \centering
    \includegraphics[width=0.975\textwidth]{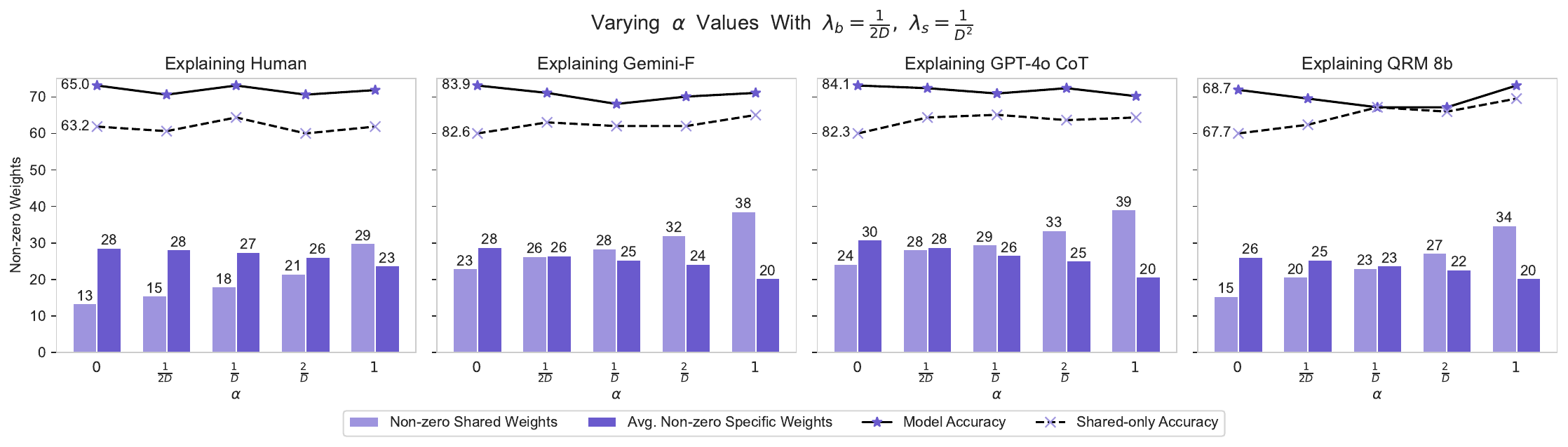}
    \includegraphics[width=0.975\textwidth]{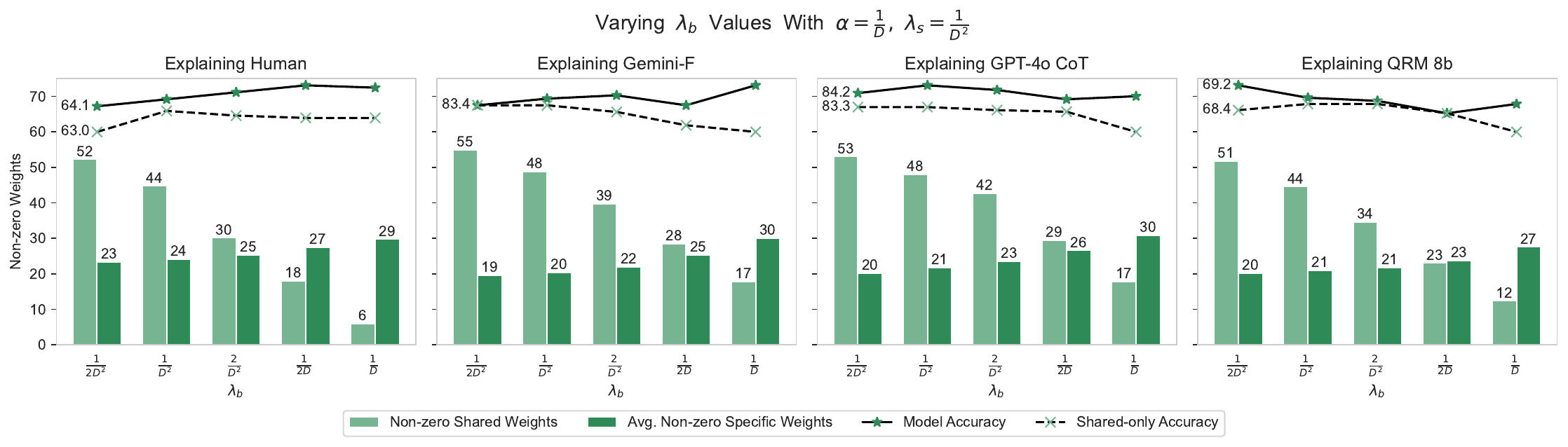}
    \includegraphics[width=0.975\textwidth]{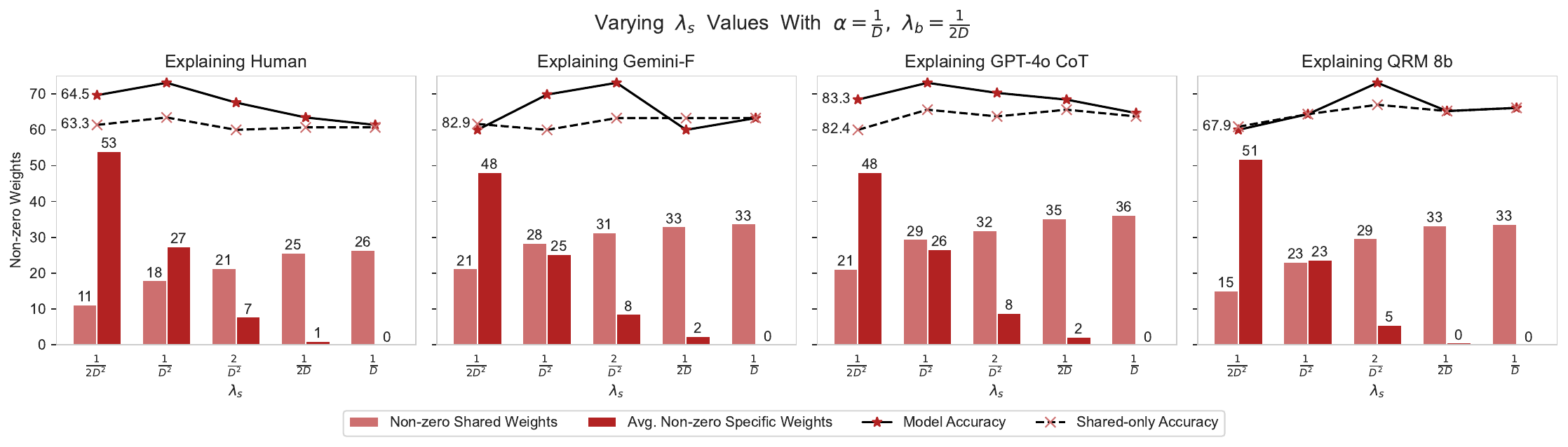}
    \caption{\textbf{Hyperparameters Analysis:} Light-colored bars represent the number of non-zero shared weights, while dark-colored bars indicate the average number of non-zero specific weights. The solid line with star markers represents model accuracy, and the dashed line with X markers shows accuracy when using only shared weights. The top figure illustrates the impact of varying $\alpha$, the middle figure examines the shared regularization parameter $\lambda_b$, and the bottom figure examines the specific regularization parameter $\lambda_s$. Results are averaged over 10 seeds.}
    \label{fig:hparams_analysis}
\vspace{-0.8em}
\end{figure*}

In the out-of-domain setup, we adopt a leave-one-out approach: in each run, one domain is excluded during training, and the model is evaluated on that held-out domain. Training is performed on examples from the remaining seven domains. Results are averaged over five random seeds, each using a subsample of 2,450 training examples, for a total of $40$ runs (5 partitions $\times$ 8 held-out domains). For both setups, we perform hyperparameter tuning using 5-fold cross-validation on the training set. Once the hyperparameters are selected, we retrain the model on the training set.

All experiments are conducted on an NVIDIA RTX 6000 with two 24GB GPUs, using the PyTorch framework and the Transformers library. We use the Adam optimizer for the white-box models and AdamW for the fine-tuned models.

\paragraph{HMDR and Ablation Models}

Our data is symmetric by definition, meaning that the concept-based representations depend on the order of the responses: the label is 1 if the first response is preferred, and -1 if the second is. Furthermore, the representation features themselves also depend on response order. To address this, after splitting the data into train and test sets, we augment each instance $(x, y)$ with its reversed form $(-x, -y)$. This augmentation helps eliminate noise and variability introduced by response ordering. When explaining LaaJs and RMs, we remove tie cases from the data.

The hyperprameters for the HMDR model are: 
\begin{align*}
    \alpha&=\frac{1}{|D|}, \\
\lambda_b&\in\Bigl\{\frac{2}{|D|^2}, \frac{1}{2|D|}, \frac{1}{|D|}\Bigr\} \\
\lambda_s&\in\Bigl\{\frac{1}{|D|^2}, \frac{2}{|D|^2}, \frac{1}{2|D|}, \frac{1}{|D|}\Bigr\} \\
&\text{and } \lambda_b \ge \lambda_s 
\end{align*}
This results in nine configurations. We chose to use only $\alpha = \frac{1}{|D|}$, as it balances the contributions of the shared and domain-specific losses to the overall objective. Additionally, this ensures a fair comparison with the ablation models, which are also evaluated using nine configurations. 

For the shared-only and specific-only variants, we use:
\begin{align*}
\lambda_{b/s}\in\Bigl\{&0.05, 0.1, 0.125, \\ &0.25, 0.5, 1.0, 1.5, 2.5, 5.0\}
\end{align*}

The dirty model we implement in this work is a variant of the original formulation by \citet{JalaliRSR10}, which decomposes domain weights into shared and domain-specific components using matrices. In our HMDR model, we modify this framework by enforcing a shared vector, instead of a shared matrix, where each row corresponds to a domain (task), which is suitable for multi-task learning. We also introduce a loss term that explicitly encourages the shared predictor to perform well independently. Accordingly, we set $\alpha = 0$ for the dirty model. Additionally, the original dirty model applies weight regularization at the row level, encouraging entire shared matrix columns to be zeroed out, rather than individual weights. The hyperparameters are identical to the ones of the HMDR model, except for $\alpha=0$.

Figure~\ref{fig:hparams_analysis} demonstrates how different hyperparameters affect model performance (Comp-rep) and the number of non-zero weights for shared and domain-specific concepts.

\paragraph{LLM-as-Judges} The prompt used for the LaaJs is shown in Box~\ref{box:llm_judge} and is based on the prompt of \citet{YeWHCZMGG0CC025}. We call the LaaJ twice, each time with a different response order. If the predictions are inconsistent, we treat the case as a tie, contributing 0.5 to the preference prediction accuracy.

\paragraph{Fine-tuned Models} 
To ensure a fair comparison, we fine-tune several NLP models on the same data. Specifically, we experiment with encoder-only models that support large context windows: BigBird-base \citep{bigbird}, LongFormer-base \citep{longformer}, and ModernBERT-base and -large \citep{modernbert}. Additionally, we fine-tune Llama-3.1-8B-Instruct and Qwen2.5-1.5B-Instruct \citep{qwen2.5} using LoRA. Reported results are the average accuracy across five train–test splits, following hyperparameter tuning of the learning rate (5e-5, 1e-5, 5e-6, 1e-6). Fine-tuning is performed using a development set of 500 examples for early stopping and learning rate selection. We use a batch size of 1 with gradient accumulation over 32 steps, a 10\% warmup ratio, and a weight decay of 0.01. Encoder-only models are trained for 20 epochs. For LoRA, we train for 5 epochs using a rank of 16, an alpha of 32, and a dropout rate of 0.05.

\subsection{Concept Importance via Lift Decomposition}
\label{app:approximating}

In this section, we show how to quantify the importance of each concept by measuring the \emph{lift} in predicted probability resulting from increasing its value by one unit. We decompose this effect into contributions from the shared weight vector $b$ and the domain-specific vector $s^{(d)}$.

Let $b_j$ and $s^{(d)}_j$ denote the shared and domain-specific weights corresponding to the $j$-th concept, whose importance we aim to quantify. Given
\[
z = (b + s^{(d)})^\top x,
\quad
\Delta z_j = b_j + s_j^{(d)}
\]
We define the \emph{lift}\footnote{In practice, we display the lift as a percentage by multiplying its value by 100.} of the $j$-th concept at input $x$ as the relative increase in predicted probability
\[
  \mathrm{lift}_j(x)
    = \frac{\Delta p^{(d)}}{p^{(d)}}
    = \frac{\sigma\bigl(z + \Delta z_j\bigr) - \sigma(z)}{\sigma(z)}
\]
\noindent Using a first-order Taylor expansion of \(\sigma\) around \(z\), we get
\begin{align*}
  \sigma\bigl(z + \Delta z_j\bigr)
  &\approx \sigma(z) + \sigma'(z)\,\Delta z_j \\[-0.25ex]
  &= \sigma(z) + \sigma(z)\bigl(1 - \sigma(z)\bigr)\,\Delta z_j
\end{align*}
where the second-order remainder
\[
  R_2 \;=\; \tfrac12\,\sigma''(\xi)\,(\Delta z_j)^2
\]
is negligible in practice since \(\lvert\Delta z_j\rvert<\tfrac14\) in our models and $\sigma''(\xi)$ is smaller than 0.1. 

Substituting into $\mathrm{lift}_j(x)$ gives
\begin{align*}
  \mathrm{lift}_j(x)
  & \approx \bigl(1 - \sigma(z)\bigr)\,\Delta z_j
\end{align*}

\noindent Taking expectations over the input distribution,
\begin{align*}
  \mathbb{E}\bigl[\mathrm{lift}_j\bigr]
  &\approx \mathbb{E}\bigl[1 - \sigma(z)\bigr] \Delta z_j \\[-0.5ex]
  &= \mathbb{E}\bigl[1 - \sigma(z)\bigr] b_j + \mathbb{E}\bigl[1 - \sigma(z)\bigr]s^{(d)}_j
\end{align*}

By construction, our data is symmetric (by swapping the locations of the first response with the second), so for every $(x,y)$ there is a corresponding $(-x,-y)$. Hence $\mathbb{E}[\sigma(z)]=0.5$, yielding
\[
  \mathbb{E}\bigl[\mathrm{lift}_j\bigr]
  \approx 0.5\;\bigl(b_j + s^{(d)}_j\bigr).
\]

\subsection{Explanation Evaluation}
\label{app:exp_eval}

To evaluate the explanations produced by our method, we propose two application-driven settings: Judge Hack and Tie Break.
In the Judge Hack setting, we first generate vanilla responses, which are then evaluated by the explained judge, using the prompt shown in Box~\ref{box:judge_hack_vanilla}. We also generate explanation-guided responses by prompting the generator LLM to consider four concepts during generation, as shown in Box~\ref{box:judge_hack_concepts}. Since the user queries we use in these settings are new, we train the HMDR model on the whole training set (excluding tie cases) using Comp-rep (which leads to better performance for the judges), with hyperparameters of: $\alpha=0.125, \lambda_b=0.125, \lambda_s=0.0625$.

In the Tie Break setting, we identify tie cases, examples where the LaaJ gives inconsistent predictions when the order of responses is swapped, using the standard LaaJ prompt in Box~\ref{box:llm_judge}. To resolve these ties, we use an explanation-guided prompt, shown in Box~\ref{box:tie_break}, that instructs the LaaJ to consider four concepts when making its preference prediction. As in the previous settings, we use Comp-rep with the same hyperparameters. However, to ensure fair evaluation, we exclude the specific tie examples we aim to resolve each time we train the HMDR model.

The explanations we use for analysis are based on Score-rep using $\alpha=0.125, \lambda_b=0.25, \lambda_s=0.0625$. We chose the Score-rep representation because it provides a more fine-grained view of concept impact, unlike the Comp-rep, which uses ternary features. We increase the value of $\lambda_s$ to put more emphasis on domain-specific effects.  

We also note that, as seen in Figure~\ref{fig:hparams_analysis}, the explanation is sensitive to hyperparameters. The hyperparameters, which control the balance between the shared loss and the domain-specific loss and the sparsity of the weights, impact the number of non-zero concepts. Even though they produce different explanations, we find the leading concepts (i.e., the concepts with the largest weight magnitude) are roughly the same. Therefore, we believe the impact on the two evaluation settings is minor in the case of using a different set of hyperparameters when training the HMDR model.


\begin{table*}[t]
\centering
\footnotesize
\begin{adjustbox}{width=0.7\textwidth}
\begin{tabular}{l|cccccccc|c}
\toprule
& \textbf{General} & \textbf{Software} &  \textbf{Legal} &   \textbf{Food} &  \textbf{Travel} &  \textbf{Picks} &   \textbf{UFB} &  \textbf{PKU} &   \textbf{Mean} \\
\midrule
Ours (Comp-rep) &     63.9 &      70.8 &   62.3 &  61.2 &    66.1 &              67.1 &           71.6 & 61.0 &  65.5 \\
Shared (Comp-rep) &     63.8 &      65.0 &   62.6 &  56.2 &    63.7 &              60.7 &           67.1 & 59.3 &  62.3 \\
Specific (Comp-rep) &     64.8 &      66.1 &   62.6 &  56.3 &    66.9 &              61.2 &           71.0 & 60.3 &  63.6 \\
Dirty (Comp-rep) &     65.0 &      71.0 &   64.3 &  58.2 &    62.6 &              67.8 &           71.1 & 60.4 &  65.1 \\
\midrule
Ours (Score-rep) &     65.9 &      63.8 &   71.2 &  68.1 &    68.0 &              61.8 &           71.3 & 58.1 &  66.0 \\
Shared (Score-rep) &     66.6 &      64.6 &   67.4 &  63.9 &    65.0 &              61.8 &           66.9 & 55.0 &  63.9 \\
Specific (Score-rep) &     65.0 &      62.0 &   69.3 &  65.2 &    60.6 &              58.4 &           73.9 & 58.5 &  64.1 \\
Dirty (Score-rep) &     64.1 &      64.6 &   69.5 &  66.3 &    65.9 &              58.8 &           72.2 & 58.2 &  65.0 \\
\midrule
GPT-4o &     64.0 &      66.1 &   63.5 &  59.4 &    59.6 &              61.0 &           73.9 & 64.6 &  64.0 \\
\null\quad +  CoT &     61.8 &      66.1 &   62.9 &  58.9 &    57.8 &              62.4 &           71.3 & 67.5 &  63.6 \\
GPT-4o-mini &     57.5 &      66.1 &   65.5 &  59.8 &    58.1 &              62.9 &           70.1 & 64.9 &  63.1 \\
\null\quad +  CoT &     60.5 &      66.2 &   66.1 &  60.4 &    57.4 &              60.3 &           69.4 & 63.1 &  62.9 \\
\midrule
Gemini-P &     63.0 &      67.1 &   64.5 &  60.9 &    59.3 &              62.3 &           73.4 & 61.5 &  64.0 \\
\null\quad +  CoT &     61.6 &      69.3 &   65.6 &  60.5 &    58.8 &              59.8 &           71.0 & 62.6 &  63.6 \\
Gemini-F &     65.6 &      68.9 &   67.3 &  61.0 &    60.6 &              62.9 &           71.0 & 59.6 &  64.6 \\
\null\quad +  CoT &     62.5 &      65.5 &   67.3 &  58.6 &    58.1 &              61.8 &           67.4 & 60.4 &  62.7 \\
\null\quad +  5-shots x8 &     64.4 &      70.7 &   67.2 &  61.2 &    62.8 &              63.9 &           72.1 & 62.7 &  65.6 \\
\null\quad +  5-shots + CoT x8 &     64.2 &      68.1 &   66.9 &  59.4 &    58.6 &              60.6 &           68.5 & 58.9 &  63.1 \\
\null\quad +  10-shots x8 &     67.0 &      72.1 &   66.2 &  61.8 &    63.4 &              64.4 &           70.1 & 63.0 &  66.0 \\
\null\quad +  10-shots + CoT x8 &     63.7 &      68.5 &   67.6 &  58.4 &    58.9 &              61.8 &           69.5 & 60.6 &  63.6 \\
\midrule
Mistral-V3 &     56.6 &      62.5 &   65.8 &  56.2 &    56.6 &              59.8 &           55.0 & 53.2 &  58.2 \\
\null\quad +  CoT &     57.1 &      60.1 &   64.0 &  56.6 &    52.2 &              59.3 &           53.6 & 54.5 &  57.2 \\
Llama-3.1 &     62.9 &      66.4 &   64.7 &  58.9 &    60.9 &              61.5 &           67.0 & 60.1 &  62.8 \\
\null\quad +  CoT &     60.8 &      66.2 &   61.4 &  58.3 &    56.4 &              62.5 &           65.1 & 60.8 &  61.4 \\
\null\quad + LoRA &     55.0 &      71.8 &   72.7 &  69.2 &    63.6 &              67.8 &           64.5 & 60.4 &  65.8 \\
FT Qwen 2.7b + LoRA &     61.5 &      55.5 &   56.5 &  55.5 &    55.2 &              58.5 &           60.0 & 56.0 &  57.3 \\
\midrule
FT ModernBERT-B &     60.6 &      61.3 &   62.7 &  57.3 &    61.4 &              57.6 &           57.0 & 59.8 &  59.7 \\
FT ModernBERT-L &     61.8 &      63.0 &   66.6 &  59.9 &    63.4 &              58.3 &           54.9 & 61.6 &  61.2 \\
  FT BigBird &     62.8 &      60.6 &   56.7 &  58.0 &    62.4 &              58.6 &           57.2 & 56.5 &  59.1 \\
FT LongFormer &     61.1 &      62.9 &   56.8 &  56.1 &    58.2 &              59.2 &           56.0 & 62.2 &  59.1 \\
\midrule
QRM 8b &     54.5 &      58.5 &   55.7 &  53.1 &    49.0 &              55.2 &           71.3 & 64.0 &  57.7 \\
Skywork 8b &     53.2 &      57.4 &   55.7 &  55.1 &    51.6 &              59.8 &           73.5 & 60.1 &  58.3 \\
\bottomrule
\end{tabular}
\end{adjustbox}
\caption{\textbf{Human Preferences -- Full Results:} The rightmost column presents the mean accuracy across the eight domains. LaaJ scores correspond to zero-shot, Chain-of-Thought (CoT), or few-shot settings, where `x8' denotes that the final prediction is a majority vote across an ensemble of eight few-shots, each with 5 or 10 demonstrations.}
\label{tab:appendix_human}
\end{table*}
\begin{table*}[t]
\centering
\footnotesize
\begin{adjustbox}{width=0.7\textwidth}
\begin{tabular}{l|cccccccc|c}
\toprule
\multicolumn{10}{c}{\textbf{In Domain Performance -- Comp-representation}} \\
\midrule
Explained Mech & \textbf{General} & \textbf{Software} &  \textbf{Legal} &   \textbf{Food} &  \textbf{Travel} &  \textbf{Picks} &   \textbf{UFB} &  \textbf{PKU} &   \textbf{Mean} \\
\midrule
Human &     63.9 &      70.8 &   62.3 &  61.2 &    66.1 &              67.1 &           71.6 & 61.0 & 65.5 \\
\midrule
GPT-4o &     79.0 &      83.7 &   63.2 &  84.7 &    84.7 &              83.9 &           86.7 & 74.1 & 80.0 \\
\null\quad + CoT &     87.8 &      87.0 &   65.6 &  86.8 &    85.4 &              85.7 &           89.4 & 78.4 & 83.2 \\
GPT-4o-mini &     78.3 &      81.4 &   63.6 &  83.6 &    84.3 &              83.0 &           84.8 & 77.8 & 79.6 \\
\null\quad + CoT &     79.5 &      82.8 &   62.6 &  85.0 &    85.4 &              86.0 &           90.0 & 79.1 & 81.3 \\
\midrule
Gemini-P &     86.7 &      88.4 &   64.8 &  88.7 &    88.6 &              86.2 &           90.2 & 79.2 & 84.1 \\
\null\quad + CoT &     91.2 &      91.5 &   65.7 &  87.4 &    90.3 &              88.7 &           87.0 & 77.8 & 85.0 \\
Gemini-F &     85.9 &      84.6 &   65.0 &  86.2 &    88.6 &              86.1 &           89.0 & 76.9 & 82.8 \\
\null\quad +10-shots x8 &     81.5 &      74.7 &   67.3 &  83.1 &    85.1 &              82.6 &           85.6 & 66.0 & 78.2 \\
\midrule
Llama-3.1 &     84.9 &      80.5 &   73.9 &  82.6 &    82.6 &              83.8 &           88.4 & 76.3 & 81.6 \\
\midrule
QRM 8b &     66.9 &      69.5 &   71.6 &  73.8 &    65.8 &              71.5 &           75.4 & 62.3 & 69.6 \\
Skywork 8b &     65.5 &      70.8 &   73.8 &  72.4 &    68.4 &              70.5 &           72.1 & 64.2 & 69.7 \\
\bottomrule
\toprule
\multicolumn{10}{c}{\textbf{In Domain Performance -- Score-representation}} \\
\midrule
Explained Mech & \textbf{General} & \textbf{Software} &  \textbf{Legal} &   \textbf{Food} &  \textbf{Travel} &  \textbf{Picks} &   \textbf{UFB} &  \textbf{PKU} &   \textbf{Mean} \\
\midrule
Human &     65.9 &      63.8 &   71.2 &  68.1 &    68.0 &              61.8 &           71.3 & 58.1 & 66.0 \\
\midrule
GPT-4o &     80.1 &      81.9 &   63.7 &  88.1 &    85.8 &              81.4 &           83.0 & 75.4 & 79.9 \\
\null\quad + CoT &     87.2 &      83.7 &   68.4 &  88.2 &    89.0 &              84.4 &           83.4 & 69.0 & 81.6 \\
GPT-4o-mini &     77.0 &      83.5 &   67.9 &  84.8 &    86.6 &              80.6 &           76.3 & 70.8 & 78.4 \\
\null\quad + CoT &     77.9 &      82.6 &   60.6 &  85.7 &    85.4 &              83.9 &           82.6 & 68.5 & 78.4 \\
\midrule
Gemini-P &     88.9 &      84.5 &   61.6 &  92.1 &    84.8 &              86.2 &           82.4 & 77.0 & 82.2 \\
\null\quad + CoT &     86.2 &      84.6 &   71.1 &  87.2 &    86.7 &              86.6 &           81.8 & 76.4 & 82.6 \\
Gemini-F &     81.4 &      84.2 &   66.4 &  86.7 &    88.4 &              84.7 &           81.9 & 74.9 & 81.1 \\
\null\quad + 10-shots x8 &     81.4 &      76.8 &   70.2 &  83.6 &    82.0 &              80.2 &           77.7 & 70.9 & 77.8 \\
\midrule
Llama-3.1 &     80.4 &      78.2 &   71.4 &  82.1 &    78.7 &              84.2 &           84.9 & 69.7 & 78.7 \\
\midrule
QRM 8b &     66.6 &      68.9 &   75.7 &  68.2 &    68.2 &              70.4 &           72.8 & 60.3 & 68.9 \\
Skywork 8b &     69.7 &      68.9 &   76.7 &  67.4 &    67.4 &              68.6 &           72.1 & 55.8 & 68.3 \\
\bottomrule
\end{tabular}
\end{adjustbox}
\caption{\textbf{Our Method -- Full In-Domain Results:} The rightmost column presents the mean accuracy across eight domains. Accuracy scores were computed using 25 train-test splits, each with 400 test instances.}
\label{tab:appendix_in_domain}
\end{table*}
\begin{table*}[t]
\centering
\footnotesize
\begin{adjustbox}{width=0.75\textwidth}
\begin{tabular}{l|cccccccc|c}
\toprule
\multicolumn{10}{c}{\textbf{Out-of-Domain Performance -- Comp-representation}} \\
\midrule
Explained Mech & \textbf{General} & \textbf{Software} &  \textbf{Legal} &   \textbf{Food} &  \textbf{Travel} &  \textbf{Picks} &   \textbf{UFB} &  \textbf{PKU} &   \textbf{Mean} \\
\midrule
Human &     64.6 &      66.8 &   61.0 &  59.1 &    63.4 &              60.4 &           67.9 & 60.5 & 62.9 \\
\midrule
GPT-4o &     78.6 &      83.0 &   63.7 &  84.8 &    84.6 &              82.6 &           85.6 & 70.1 & 79.1 \\
\null\quad + CoT &     86.8 &      86.4 &   69.0 &  88.4 &    86.8 &              84.7 &           89.5 & 68.8 & 82.6 \\
GPT-4o-mini &     79.3 &      82.4 &   64.7 &  81.7 &    84.8 &              83.1 &           85.4 & 72.7 & 79.3 \\
\null\quad + CoT &     80.3 &      82.8 &   63.5 &  84.4 &    83.9 &              84.4 &           89.5 & 71.8 & 80.1 \\
\midrule
Gemini-P &     87.1 &      88.4 &   67.3 &  88.5 &    87.3 &              85.2 &           89.5 & 73.7 & 83.4 \\
\null\quad + CoT &     86.5 &      87.3 &   69.0 &  87.7 &    87.9 &              87.7 &           85.7 & 68.0 & 82.5 \\
Gemini-F &     84.0 &      83.0 &   67.2 &  86.2 &    87.9 &              85.2 &           87.3 & 71.9 & 81.6 \\
\null\quad + 10-shots x8 &     81.4 &      78.4 &   66.8 &  82.4 &    82.4 &              82.5 &           84.1 & 66.6 & 78.1 \\
\midrule
Llama-3.1 &     81.5 &      76.8 &   71.3 &  82.4 &    83.4 &              83.9 &           87.0 & 67.2 & 79.2 \\
\midrule
QRM 8b &     65.9 &      69.2 &   72.5 &  73.6 &    66.9 &              68.2 &           75.4 & 59.6 & 68.9 \\
Skywork 8b &     68.0 &      69.2 &   75.7 &  72.4 &    66.6 &              68.8 &           74.1 & 58.3 & 69.1 \\
\bottomrule
\toprule
\multicolumn{10}{c}{\textbf{Out-of-Domain Performance -- Score-representation}} \\
\midrule
Explained Mech & \textbf{General} & \textbf{Software} &  \textbf{Legal} &   \textbf{Food} &  \textbf{Travel} &  \textbf{Picks} &   \textbf{UFB} &  \textbf{PKU} &   \textbf{Mean} \\
\midrule
Human &     66.1 &      64.9 &   66.8 &  61.9 &    65.2 &              59.8 &           69.1 & 52.7 & 63.3 \\
\midrule
GPT-4o &     79.9 &      81.5 &   67.4 &  86.6 &    85.9 &              80.6 &           81.6 & 71.5 & 79.4 \\
\null\quad + CoT &     86.5 &      84.1 &   71.1 &  89.2 &    88.6 &              83.4 &           83.5 & 67.1 & 81.7 \\
GPT-4o-mini &     76.9 &      83.9 &   68.3 &  84.2 &    85.9 &              81.2 &           76.2 & 67.4 & 78.0 \\
\null\quad + CoT &     79.9 &      82.2 &   67.6 &  85.7 &    84.9 &              84.8 &           80.2 & 66.2 & 78.9 \\
\midrule
Gemini-P &     86.7 &      86.4 &   70.6 &  90.9 &    85.4 &              84.4 &           81.0 & 73.1 & 82.3 \\
\null\quad + CoT &     86.2 &      86.7 &   71.7 &  88.3 &    84.0 &              86.2 &           80.9 & 72.6 & 82.1 \\
Gemini-F &     83.4 &      83.8 &   69.4 &  87.3 &    88.2 &              85.2 &           80.6 & 74.5 & 81.5 \\
\null\quad + 10-shots x8 &     82.1 &      77.9 &   71.1 &  83.1 &    80.4 &              78.8 &           78.9 & 71.1 & 77.9 \\
\midrule
Llama-3.1 &     80.5 &      77.3 &   73.8 &  81.3 &    80.6 &              84.4 &           84.2 & 68.0 & 78.8 \\
\midrule
QRM 8b &     66.6 &      69.7 &   76.6 &  70.9 &    71.2 &              69.4 &           71.1 & 59.6 & 69.4 \\
Skywork 8b &     68.6 &      69.2 &   77.4 &  68.4 &    68.9 &              67.3 &           70.8 & 55.3 & 68.2 \\
\bottomrule
\end{tabular}
\end{adjustbox}
\caption{\textbf{Our Method -- Full Out-of-Domain Results:} The rightmost column presents the mean out-of-domain accuracy across eight domains. Our method was trained and evaluated with one domain held out at a time. The reported scores reflect performance on the held-out domain, using only the shared weights for predictions. Accuracy scores were computed using 5 bootstraps, each with 400 test instances.}
\label{tab:appendix_ood}
\end{table*}

\section{Explanations}
\label{app:explanations}
\begin{figure*}[t]
    \centering
    \includegraphics[width=0.95\textwidth]{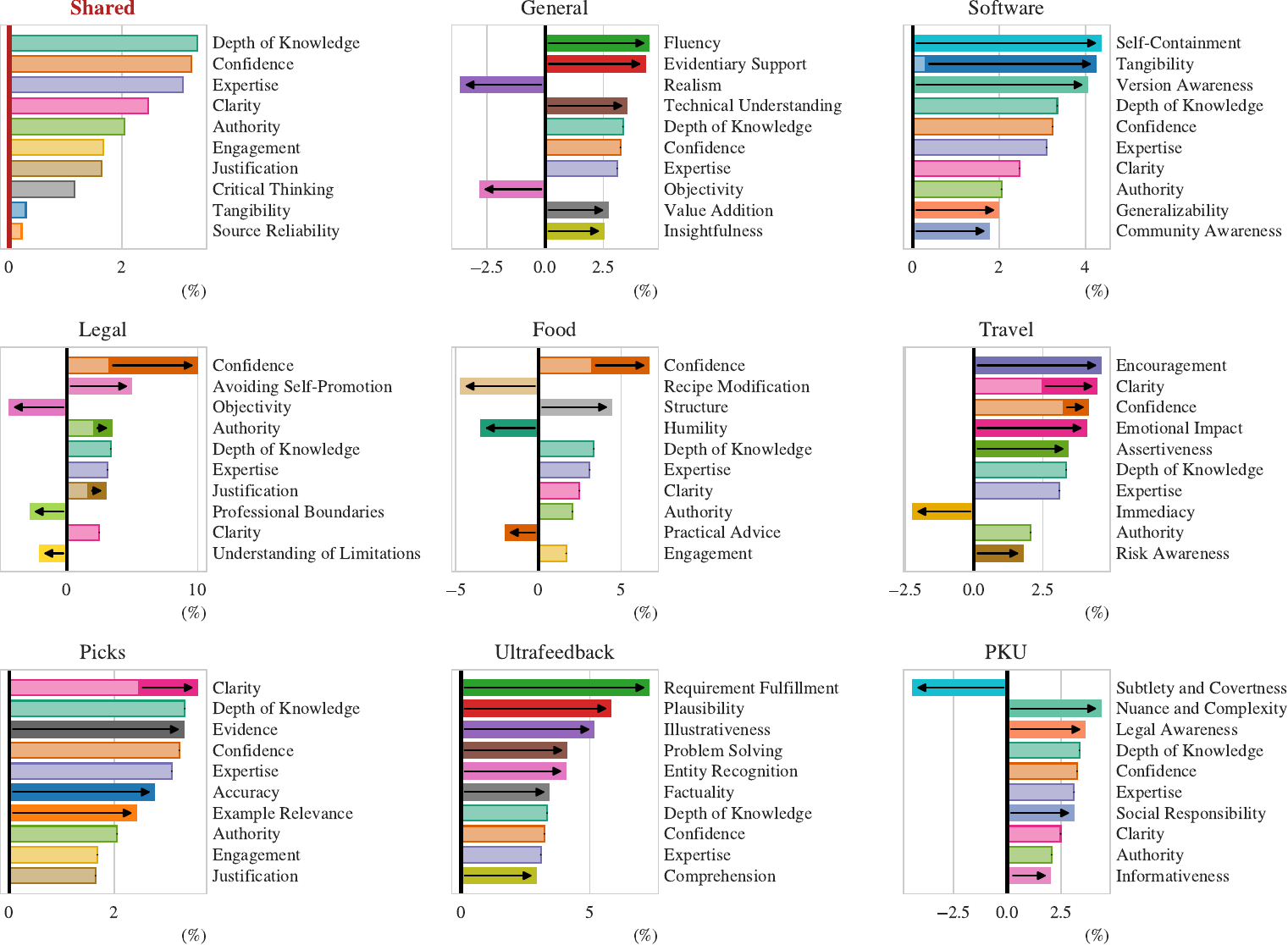}
    \vspace{-0.5em}
    \caption{\textbf{Explanations of Human Preferences.}}
    \label{fig:exp_human}
\end{figure*}
\begin{figure*}[t]
    \centering
    \includegraphics[width=0.95\textwidth]{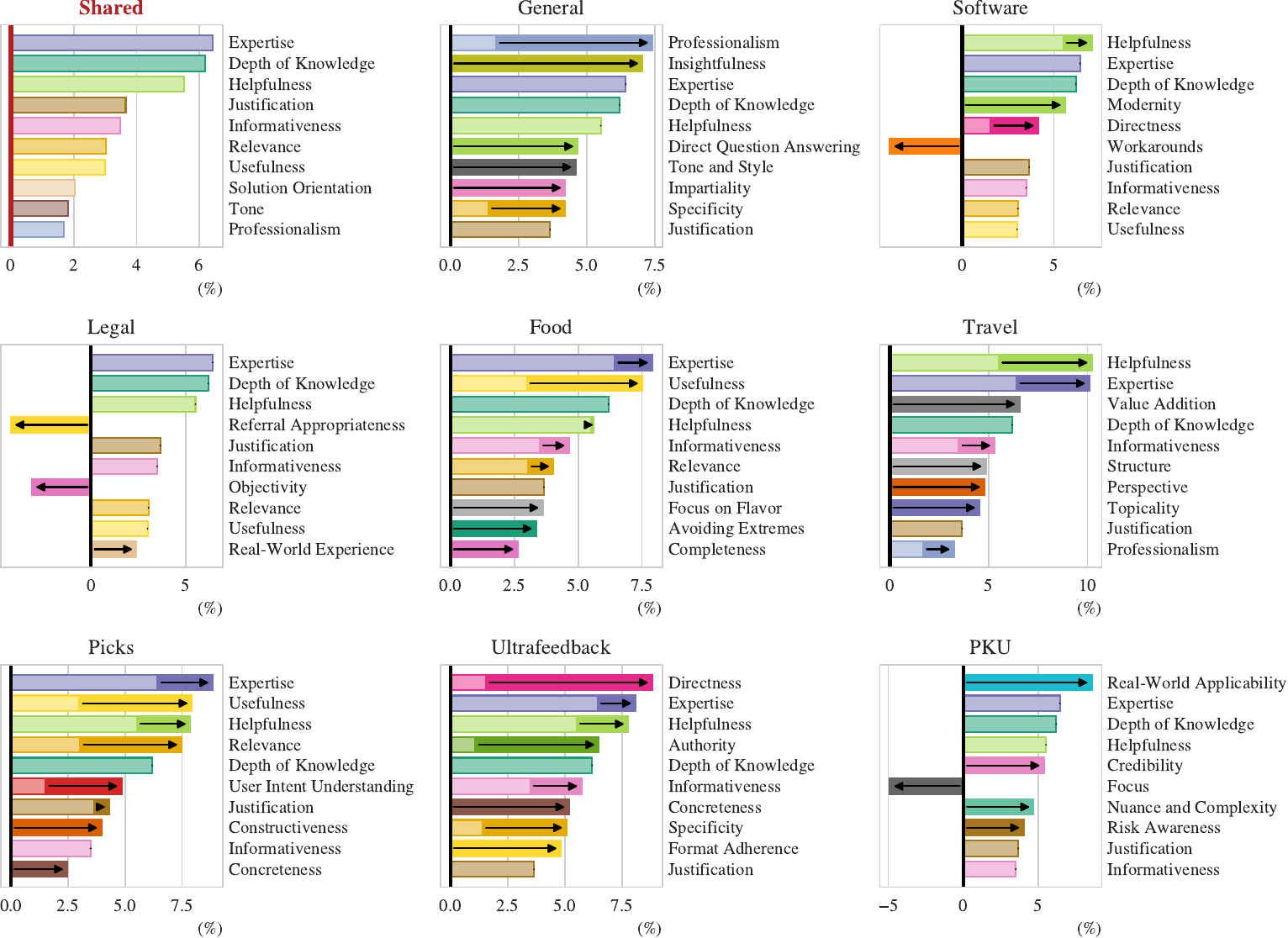}
    \vspace{-0.5em}
    \caption{\textbf{Explanations of Llama-3.1 Preferences.}}
    \label{fig:exp_llama}
\end{figure*}

\begin{figure*}[t]
    \centering
    \includegraphics[width=0.95\textwidth]{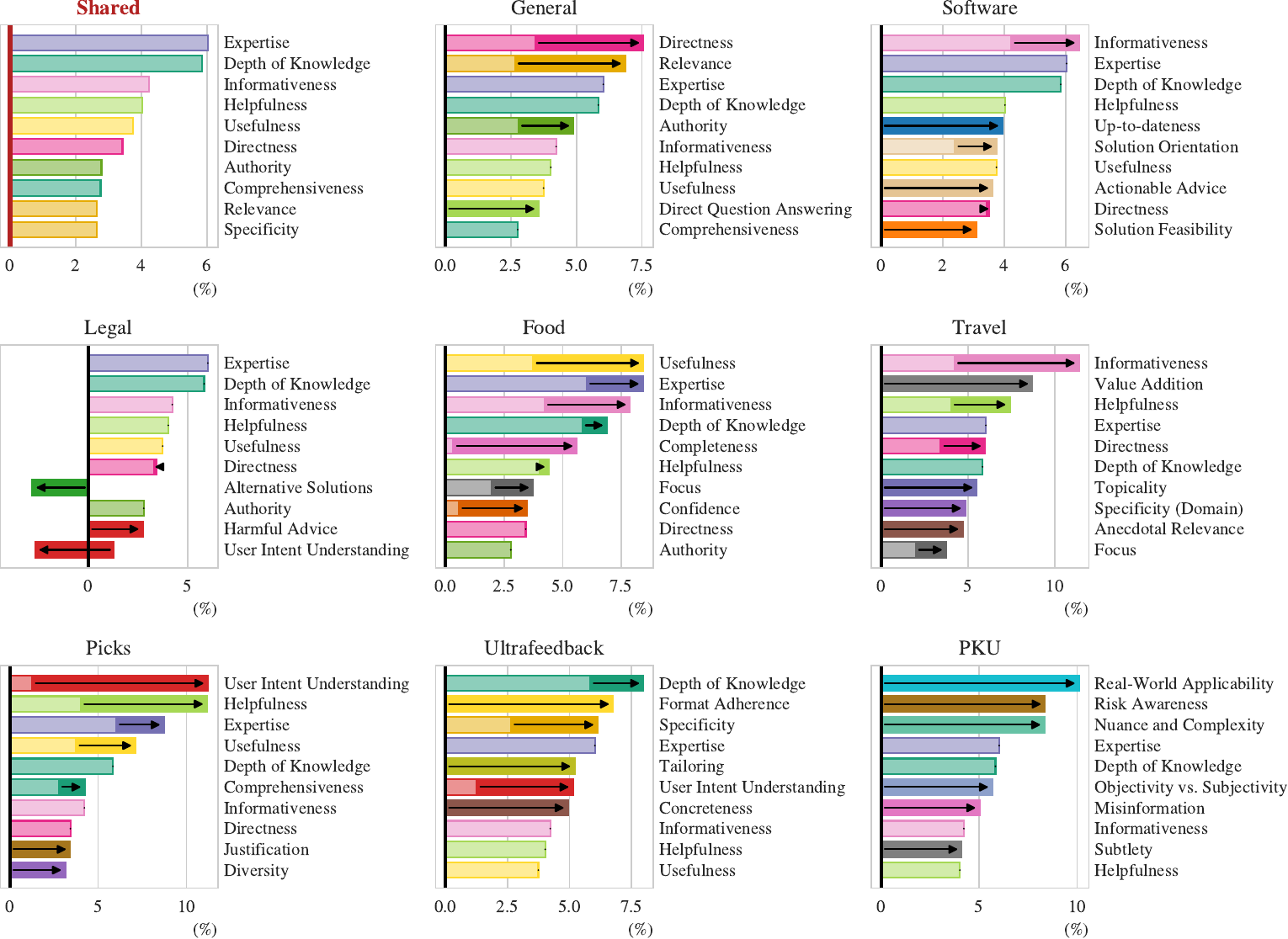}
    \vspace{-0.5em}
    \caption{\textbf{Explanations of Gemini-1.5-Flash Preferences.}}
    \label{fig:exp_gemini_f}
\end{figure*}
\begin{figure*}[t]
    \centering
    \includegraphics[width=0.95\textwidth]{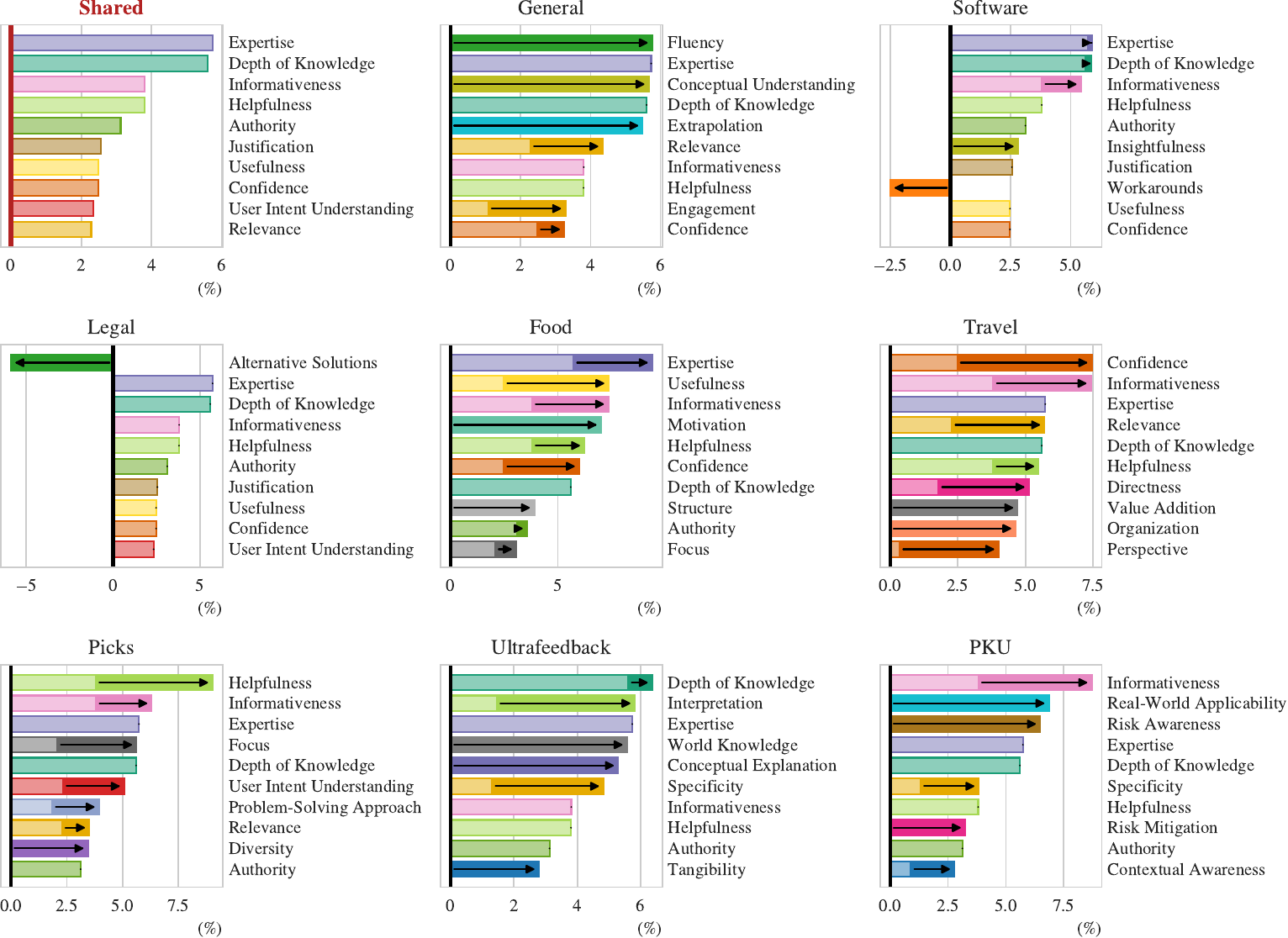}
    \vspace{-0.5em}
    \caption{\textbf{Explanations of Gemini-1.5-Flash with 10-shots Preferences.}}
    \label{fig:exp_gemini_f_10shots}
\end{figure*}
\begin{figure*}[t]
    \centering
    \includegraphics[width=0.95\textwidth]{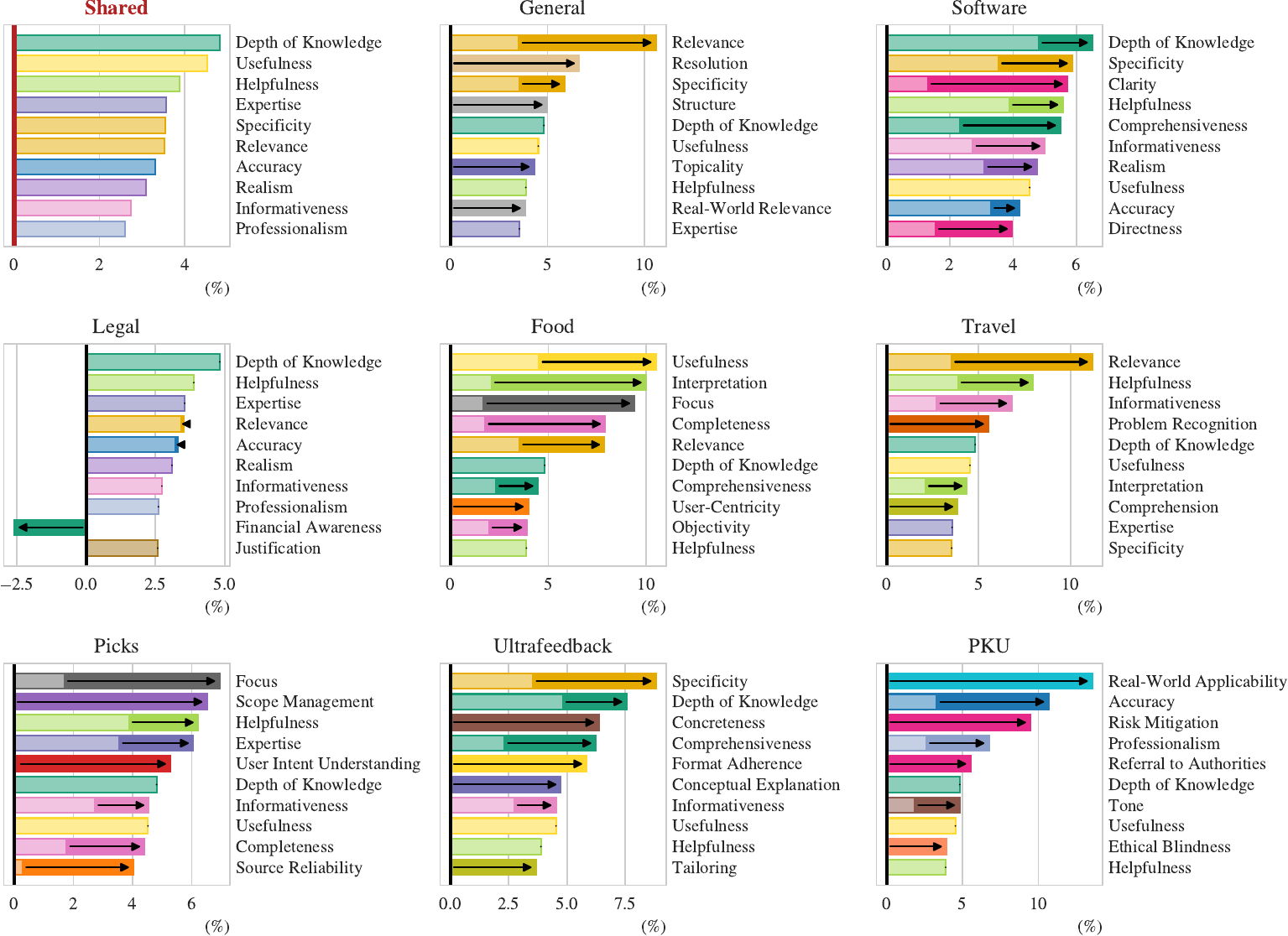}
    \vspace{-0.5em}
    \caption{\textbf{Explanations of Gemini-1.5-Pro Preferences.}}
    \label{fig:exp_gemini_p}
\end{figure*}

\begin{figure*}[t]
    \centering
    \includegraphics[width=0.95\textwidth]{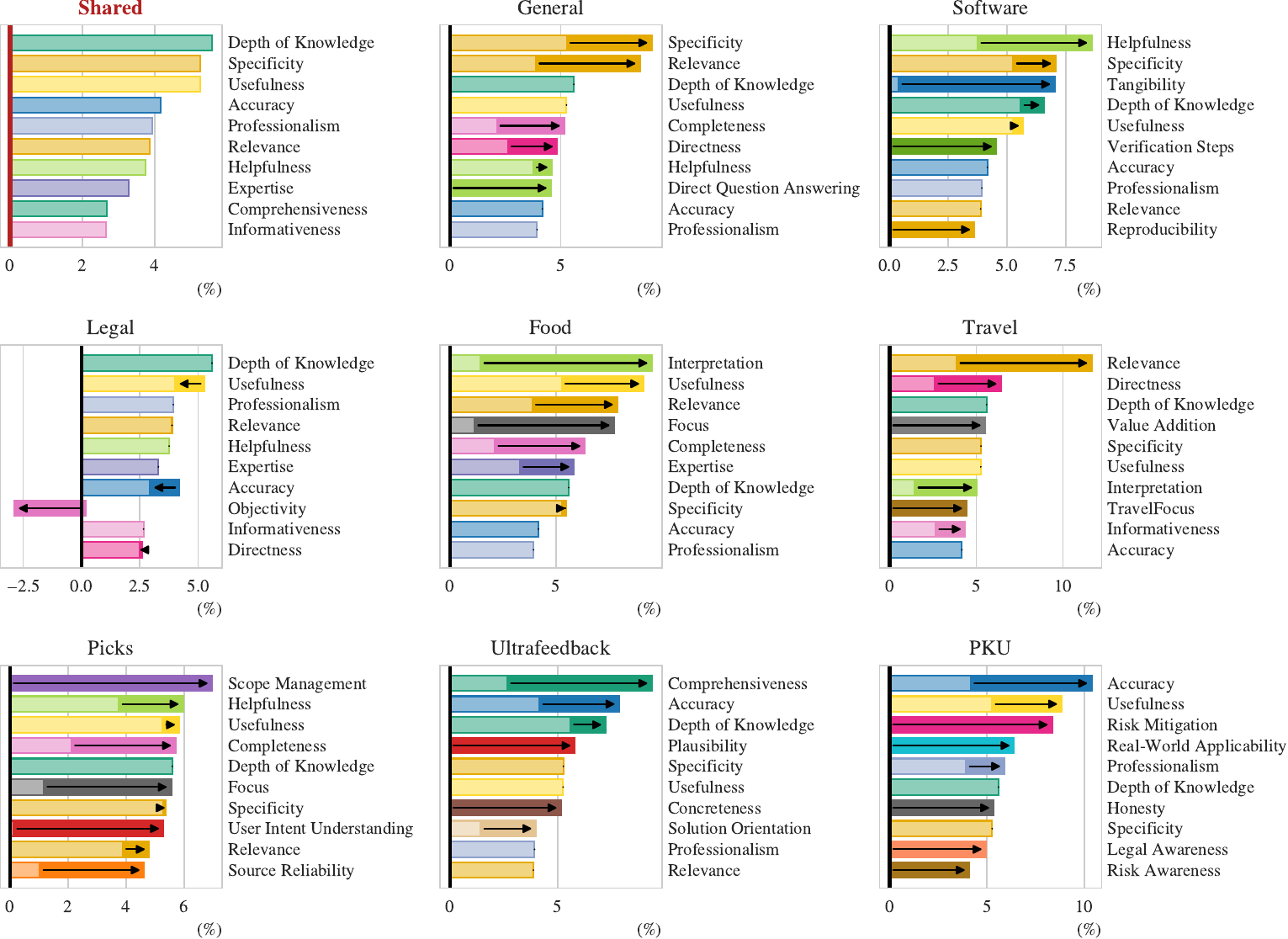}
    \vspace{-0.5em}
    \caption{\textbf{Explanations of Gemini-1.5-Pro with Chain-of-Thoughts Preferences.}}
    \label{fig:exp_gemini_p_cot}
\end{figure*}

\begin{figure*}[t]
    \centering
    \includegraphics[width=0.95\textwidth]{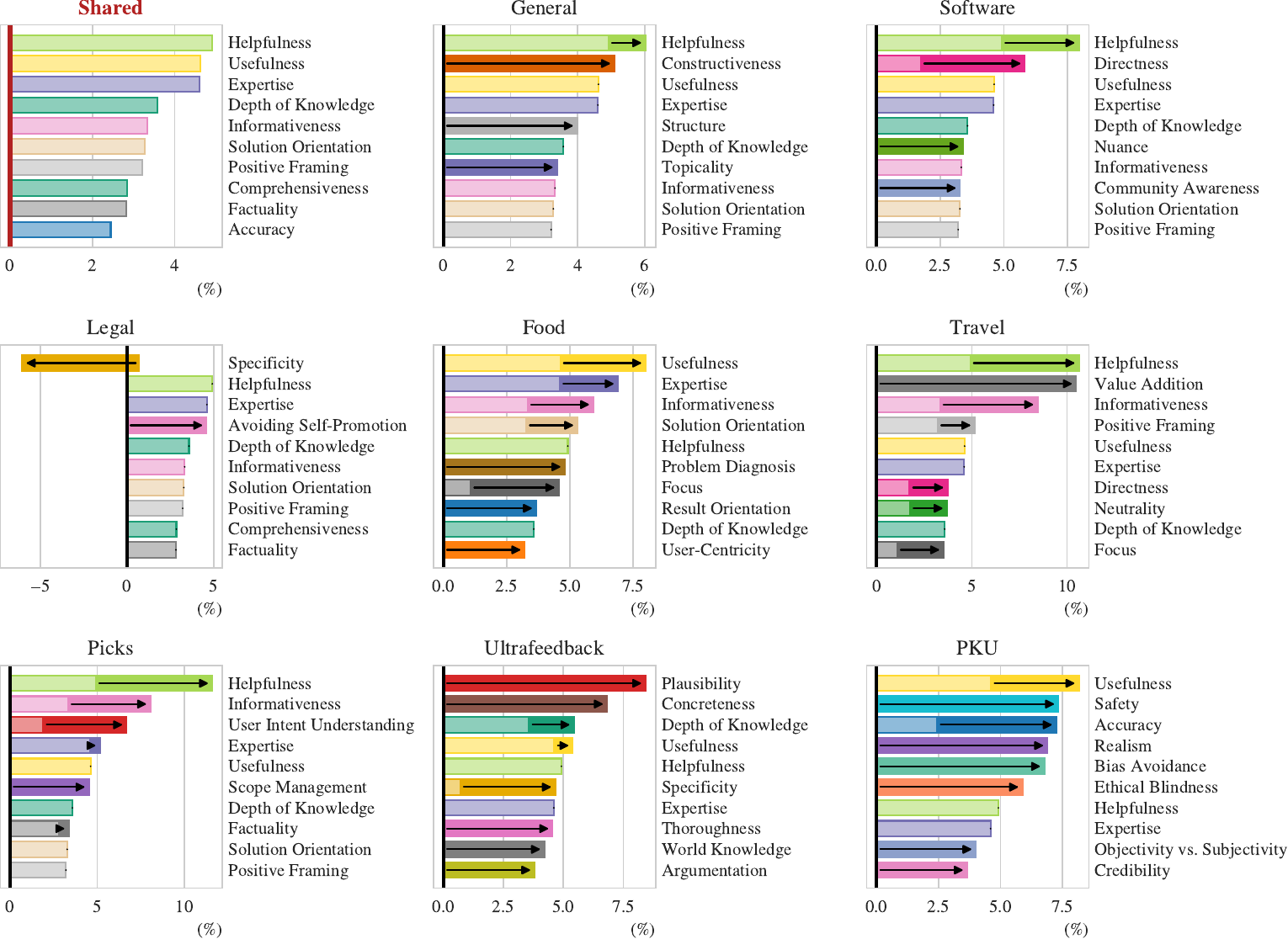}
    \vspace{-0.5em}
    \caption{\textbf{Explanations of GPT-4o Preferences.}}
    \label{fig:exp_gpt4o}
\end{figure*}

\begin{figure*}[t]
    \centering
    \includegraphics[width=0.95\textwidth]{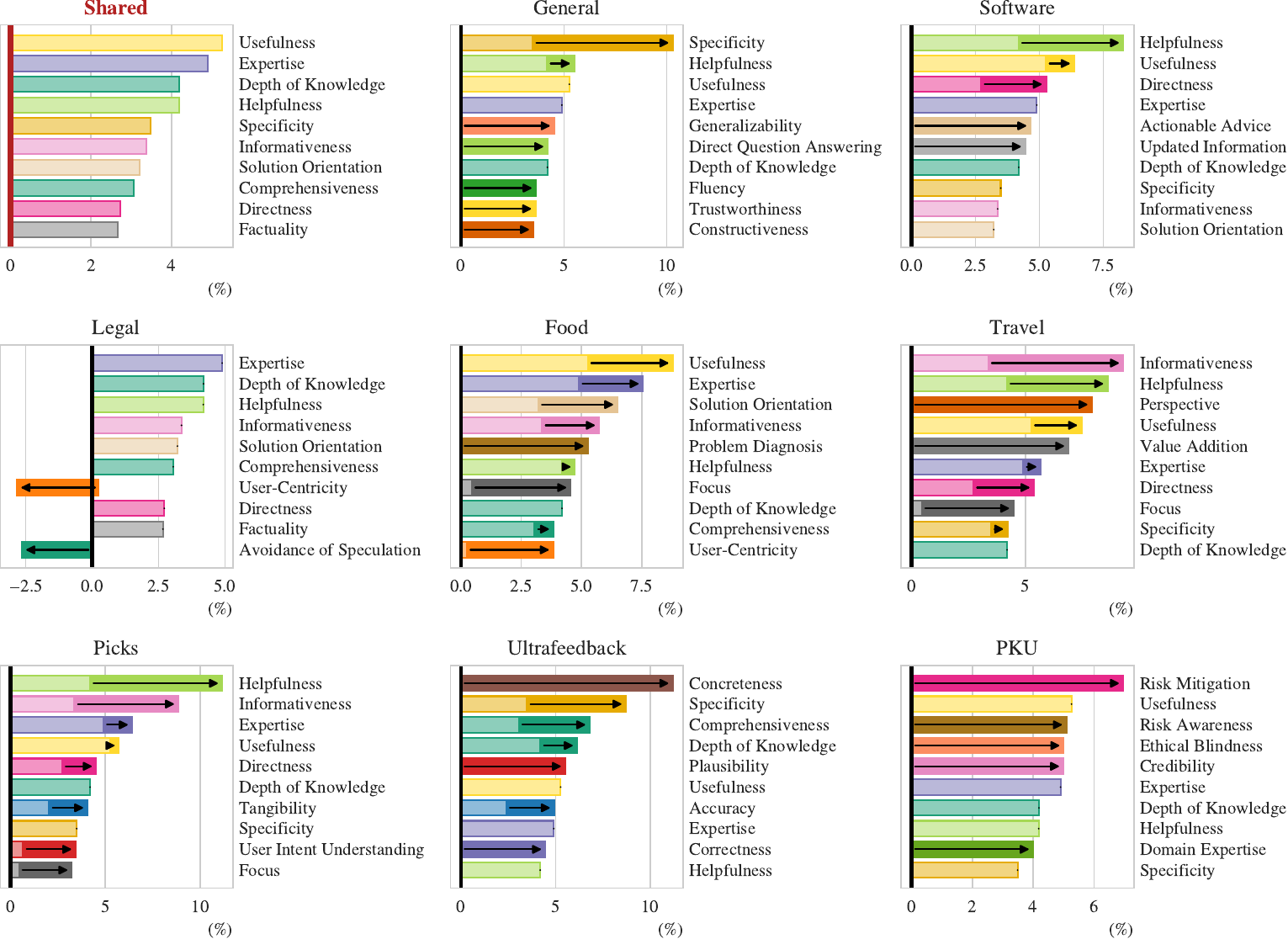}
    \vspace{-0.5em}
    \caption{\textbf{Explanations of GPT-4o with Chain-of-Thoughts Preferences.}}
    \label{fig:exp_gpt4o_cot}
\end{figure*}

\begin{figure*}[t]
    \centering
    \includegraphics[width=0.95\textwidth]{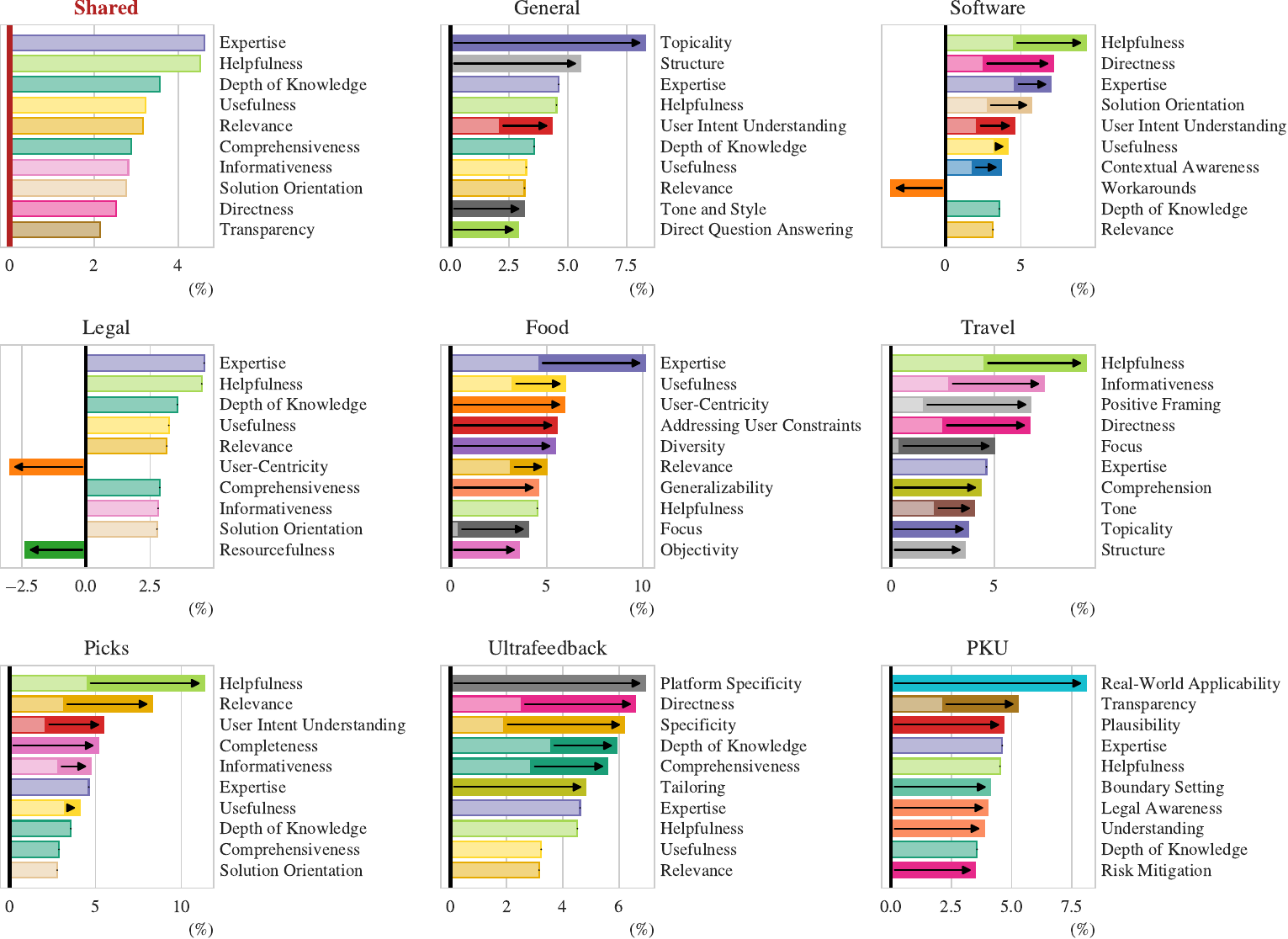}
    \vspace{-0.5em}
    \caption{\textbf{Explanations of GPT-4o-mini Preferences.}}
    \label{fig:exp_gpt4o_mini}
\end{figure*}

\begin{figure*}[t]
    \centering
    \includegraphics[width=0.95\textwidth]{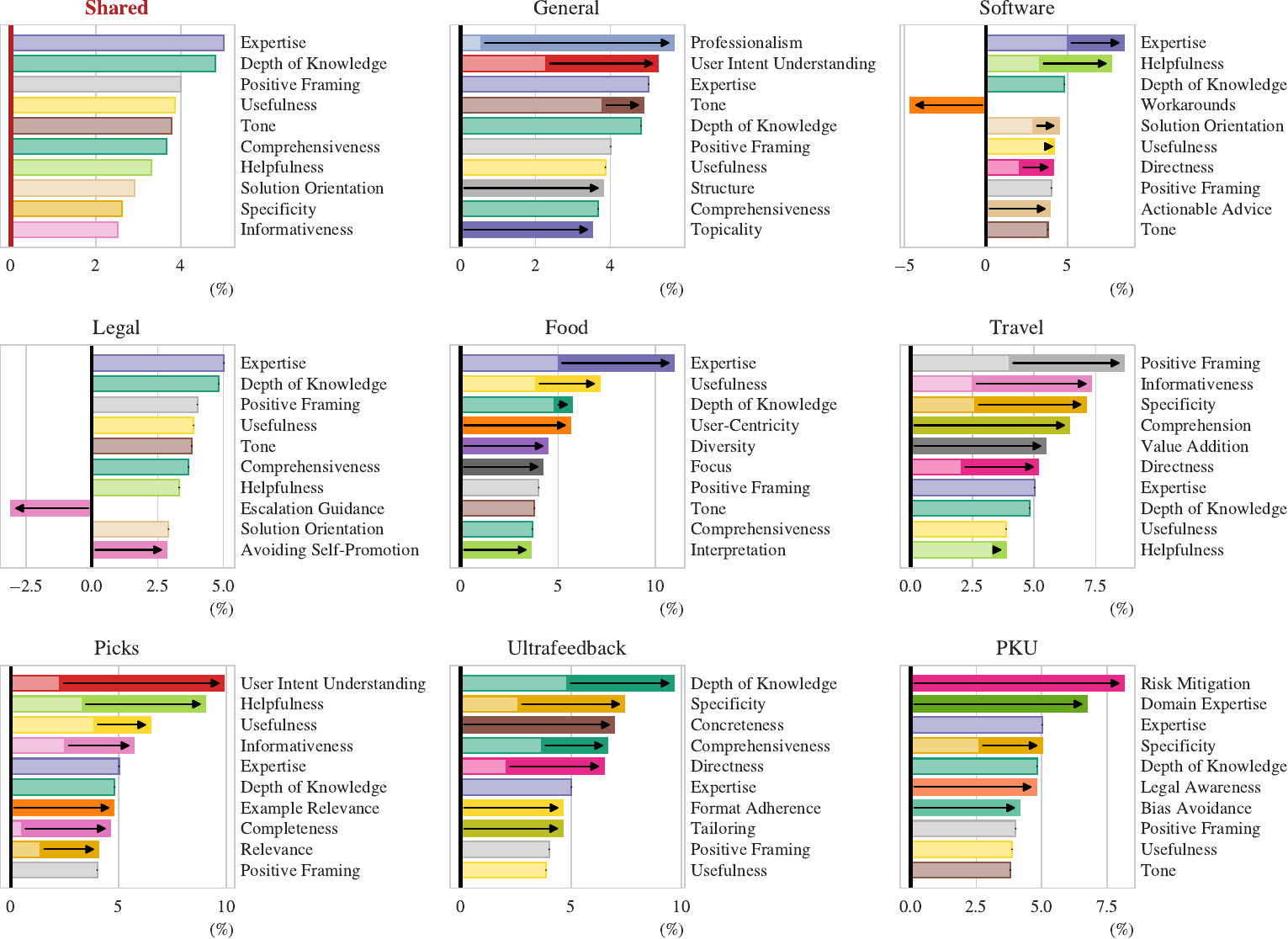}
    \vspace{-0.5em}
    \caption{\textbf{Explanations of GPT-4o-mini with Chain-of-Thoughts Preferences.}}
    \label{fig:exp_gpt4o_mini_cot}
\end{figure*}

\begin{figure*}[t]
    \centering
    \includegraphics[width=0.95\textwidth]{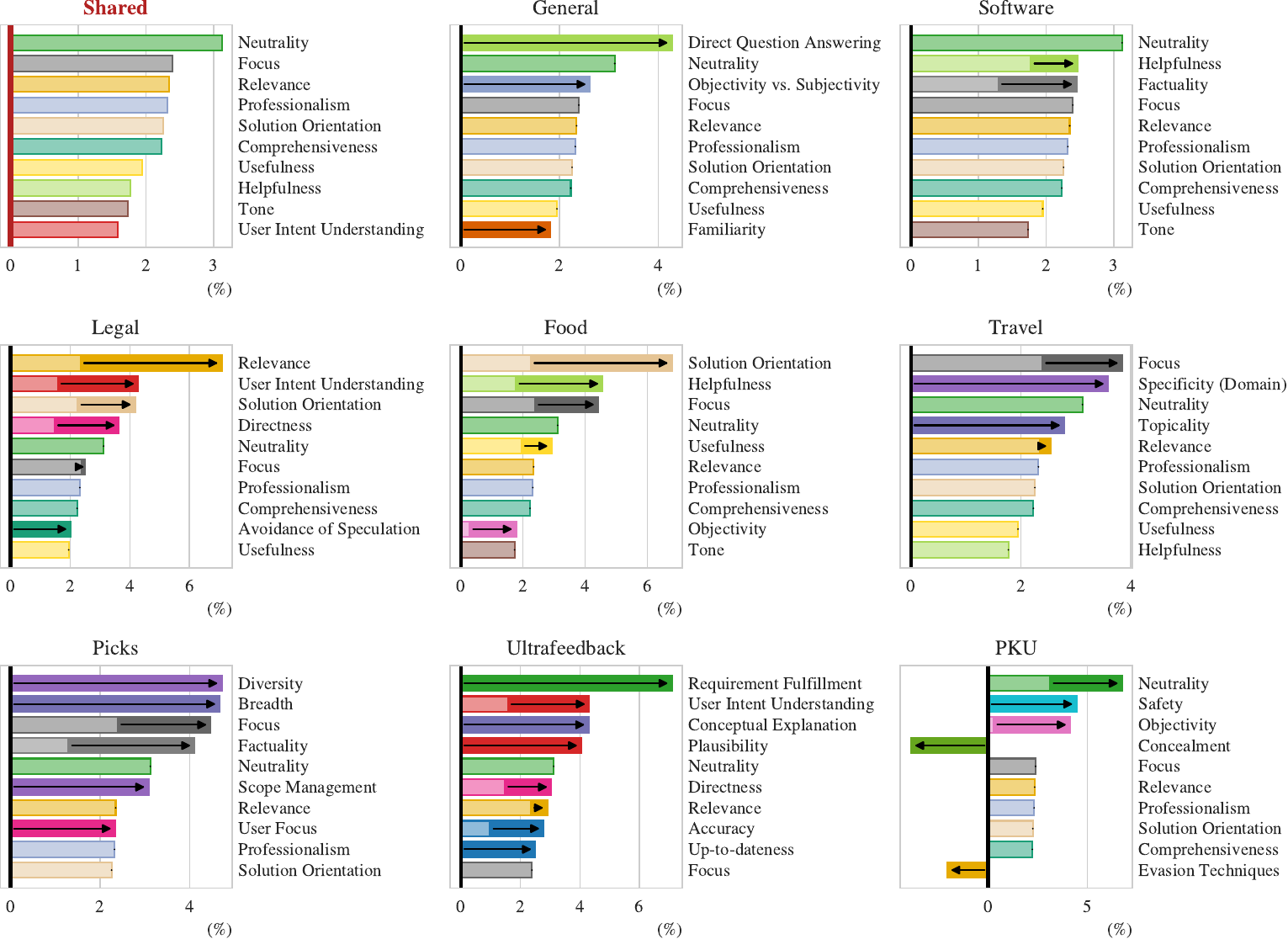}
    \vspace{-0.5em}
    \caption{\textbf{Explanations of QRM 8b Reward Model Preferences.}}
    \label{fig:exp_qrm}
\end{figure*}

\begin{figure*}[t]
    \centering
    \includegraphics[width=0.95\textwidth]{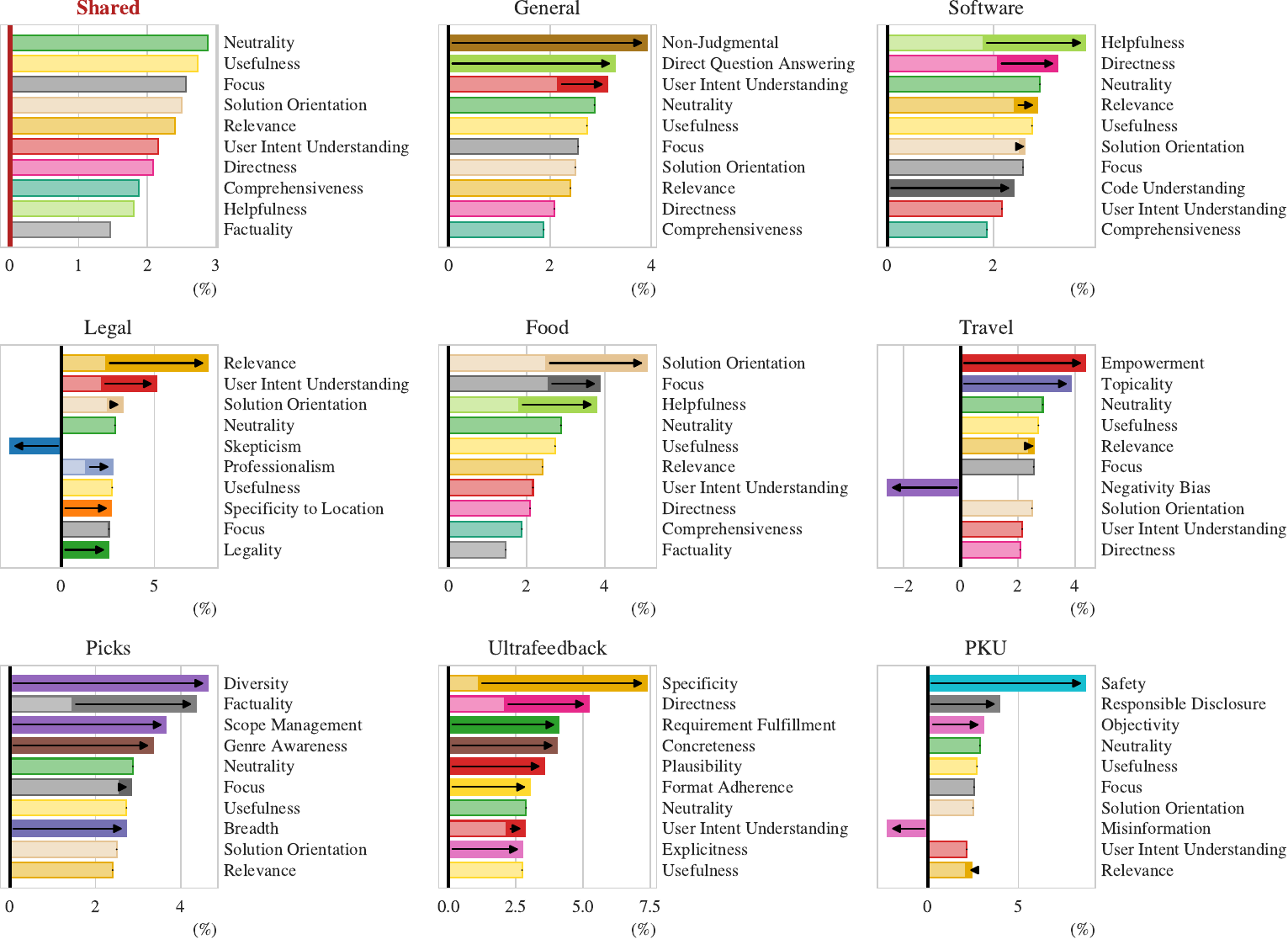}
    \vspace{-0.5em}
    \caption{\textbf{Explanations of SkyWork 8b Reward Model Preferences.}}
    \label{fig:exp_skywork}
\end{figure*}

\onecolumn
\section{Prompts}
\label{app:prompts}
\vspace{-10em}
\begin{prompt}[label={box:prop_subdomains}]{LimeGreen}{Proposing Subdomains and Tasks}
You will be provided with $n_b$ user queries. \\
Each user query may include an instruction (defining a task or the expected type of response) along with additional text.\\
Your task is to determine the domains and task types conveyed in the user queries.\\
A domain can be a general category of knowledge or a field of expertise (e.g., "healthcare", "technology", "python", ...).\\
A task type can be a specific type of task that the response is expected to address (e.g., "summarization", "translation", "question answering", ...).\\
Please provide your answers in the JSON format below, where the keys are 'domains' and 'tasks' and the values are lists of relevant domains and task types.\\
\\
{[EXAMPLES]}\\
\\
\textasciigrave\textasciigrave\textasciigrave json\\
\{\\
\null\quad "domains": ["list", "of", "relevant", "domains"],\\
\null\quad "tasks": ["list", "of", "relevant", "task", "types"]\\
\}\\
\textasciigrave\textasciigrave\textasciigrave
\end{prompt}
\vspace{-12em}
\begin{prompt}[label={box:ann_subdomains}]{LimeGreen}{Annotating Subdomains and Tasks}
You will be provided with a user query. \\
The user query may include an instruction or a question (defining the task that the response to the query is expected to address) along with additional text. \\
Your tasks are: \\
(1) Determine the domains of the user query. Select domains from the `Domains` list below or write 'None' if none of the domains are relevant to the user query. \\
(2) Determine the task types of the instruction. Select task types from the `Tasks` list below or write 'None' if none of the task types are relevant to the instruction. \\
\\
Domains:\\
{[DOMAINS]}\\
\\
Tasks:\\
{[TASKS]}\\
\\
User Query:\\
{[USER QUERY]}\\
\\
\textasciigrave\textasciigrave\textasciigrave json\\
\{\\
\null\quad "domains": ["list", "of", "relevant", "domains"],\\
\null\quad "tasks": ["list", "of", "relevant", "task", "types"]\\
\}\\
\textasciigrave\textasciigrave\textasciigrave
\end{prompt}
\vspace{-12em}
\begin{prompt}[label={box:discovery}]{Green}{Discovering Concepts}
You will be provided with $n_b$ examples, each example consists of \{a user query and two responses. One of the responses was chosen by the user, and the other was rejected | a user query and a response that was chosen by the user | a user query and a response that was rejected by the user\}. \\
Your task is to identify and describe $n_c$ concepts (or features) that can explain why the user 
\{preferred the chosen response over the rejected response | chose the response | rejected the response\}.\\
\\
Notice that the subdomain of all examples is {[SUBDOMAIN]}, and the NLP task conveyed in all user queries is {[TASK]}. 
Try to suggest domain-specific and task-specific concepts relevant to the provided examples.\\
\\
The name of each concept should be precise and clearly defined, without combining two concepts (e.g., instead of "Structure and Organization" use "Structure"). \\
In addition, write a concise one-sentence description of each concept. \\
\{\{\{Please, suggest concepts that are *completely different* from the following:\\
{[FIXED CONCEPTS]}\}\}\}\\
\\
{[BATCH]}\\
\\
Remember, your task is to identify $n_c$ concepts that can explain why the user \{preferred the chosen response over the rejected response | chose the response | rejected the response\}. \\
\\
Each concept name should start with a capital letter, the definition should be a concise one-sentence description starting with "A good response..." or "A bad response..." depending on the concept. \\
Please provide your answers in JSON format, where the keys are the concepts and the values are their definitions.\\
\\
For example:\\
\textasciigrave\textasciigrave\textasciigrave json\\
\{\\
\null\quad "Concept name": "a concise one-sentence description starting with 'A good response...' or 'A bad response...'"\\
\}\\
\textasciigrave\textasciigrave\textasciigrave
\end{prompt}

\begin{prompt}[label={box:fixed_concepts}]{Green}{Fixed Concepts}
\{\\
\null\quad  "Specificity": "A high score indicates the response provides detailed, precise information; A low score indicates the response is vague or overly general.",\\
\null\quad "Clarity": "A high score indicates the response is easy to understand and well-structured; A low score indicates the response is confusing or poorly organized.",\\
\null\quad "Relevance": "A high score indicates the response is directly related to the query or context; A low score indicates the response is off-topic or irrelevant.",\\
\null\quad "Helpfulness": "A high score indicates the response is beneficial and actionable for the user; A low score indicates the response is unhelpful or impractical.",\\
\null\quad "Empathy": "A high score indicates the response shows understanding and consideration of the user’s emotions; A low score indicates the response is indifferent or dismissive.",\\
\null\quad "Accuracy": "A high score indicates the response contains correct and factual information; A low score indicates the response is inaccurate or misleading.",\\
\null\quad  "Informativeness": "A high score indicates the response provides valuable and comprehensive information; A low score indicates the response lacks substance or detail.",\\
\null\quad "Creativity": "A high score indicates the response is original and imaginative; A low score indicates the response is unoriginal or conventional.",\\
\null\quad "Safety": "A high score indicates the response avoids harmful content and adheres to ethical standards; A low score indicates the response may contain dangerous or unethical content.",\\
\null\quad "Engagement": "A high score indicates the response captures and retains the user’s interest; A low score indicates the response is dull or unengaging."\\
\}\\
\end{prompt}

\begin{prompt}[label={box:filter}]{ForestGreen}{Semantical Duplicates}
You will be provided with pairs of concepts used to evaluate responses written by humans or generated by an LLM, along with their definitions. Each pair is a key in a dictionary. For each pair, you should determine if the two concepts are semantically identical and assess the same aspects of the response, i.e., the score of any response would be the same for both concepts.\\
\\
Fill the following JSON dict with True/False values:\\
\textasciigrave\textasciigrave\textasciigrave json\\
{[FLAGGED CONCEPT PAIRS]}\\
\textasciigrave\textasciigrave\textasciigrave
\end{prompt}

\begin{prompt}[label={box:defining}]{PineGreen}{Defining Concepts}
You will be provided with a dict where the keys are concepts used to evaluate responses written by humans or generated by an LLM. \\
Each concept includes a list of descriptions explaining the relevancy of the concept. \\
Your task is to write concise two-sentence concept definitions, each definition should be based on the descriptions and start with "A high score indicates...; A low score indicates...".\\
Please follow the JSON format below, the key is a concept, and you need to write its corresponding definition.\\
\\
Concept Descriptions:\\
{[DESCRIPTIONS]}\\
\\
Fill in the JSON format below with the definitions:\\
\textasciigrave\textasciigrave\textasciigrave json\\
{[CONCEPTS]}\\
\textasciigrave\textasciigrave\textasciigrave
\end{prompt}

\begin{prompt}[label={box:relevant}]{Peach}{Predicting Relevant Concepts}
You will be provided with a list of concepts used for evaluating responses written by humans or generated by an LLM. \\
In addition, you will be provided with a user query and two responses. Your task is to predict whether each concept is relevant for evaluating these responses. \\
Please follow the JSON format below, the key is a concept, and you need to fill in True or False according to the relevance of the concept. \\
\\
User Query:\\
{[USER QUERY]}\\
\\
Response 1: \\
{[RESPONSE 1]}\\
\\
Response 2: \\
{[RESPONSE 2]}\\
\\
Fill in the JSON format below with True or False if the concept is relevant for evaluating the responses:\\
\textasciigrave\textasciigrave\textasciigrave json\\
{[CONCEPTS]}\\
\textasciigrave\textasciigrave\textasciigrave
\end{prompt}

\begin{prompt}[label={box:comp_rep}]{BurntOrange}{Annotating Comp-rep}
You will be provided with a list of concepts used for evaluating responses written by humans or generated by an LLM. \\
In addition, you will be provided with a user query and two responses. \\
Your task is to compare the two responses and, for each concept, determine which response should be scored higher based on the concept's definition. Before determining which response should be scored higher for each concept, you must provide a concise explanation.
Conclude the explanation with "Final answer: 1" if the first response should be scored higher, "Final answer: 2" if the second response should be scored higher, or "Final answer: 0" if both responses should be scored equally or the concept is not relevant. \\
Remember to be critical and objective in your evaluation. \\
Please follow the JSON format below, the key is a concept, and you need to write an explanation and the final answer. \\
\\
Concepts: \\
{[CONCEPT DEFINITIONS]}\\
\\
User Query:\\
{[USER QUERY]}\\
\\
Response 1: \\
{[RESPONSE 1]}\\
\\
Response 2: \\
{[RESPONSE 2]}\\
\\
Fill in the JSON format below with an explanation and the final answer (0, 1, or 2) for each concept: \\
\textasciigrave\textasciigrave\textasciigrave json\\
{[CONCEPTS]}\\
\textasciigrave\textasciigrave\textasciigrave
\end{prompt}

\begin{prompt}[label={box:score_rep}]{RedOrange}{Annotating Score-rep}
You will be provided with a list of concepts used for evaluating responses written by humans or generated by an LLM. \\
In addition, you will be provided with a user query and a response. \\
Your task is to score the response according to each concept definition.
The score should be on a scale of 1 to 7, where 1 indicates the concept's score of the response is very low and 7 indicates the concept's score is very high.
Use 0 if the concept is not relevant for evaluating the response. \\
Before determining the score for each concept, you must provide a concise explanation. Conclude the explanation with "Final answer: X",  where X is the concept's score. \\
Remember to be critical and objective in your evaluation. \\
Please follow the JSON format below, the key is a concept, and you need to write an explanation and the final answer. \\
\\
Concepts: \\
{[CONCEPT DEFINITIONS]}\\
\\
User Query:\\
{[USER QUERY]}\\
\\
Response 1: \\
{[RESPONSE 1]}\\
\\
Response 2: \\
{[RESPONSE 2]}\\
\\
Fill in the JSON format below with an explanation (be critical) and the final answer (int between 1 and 7, and 0 if the concept is not relevant) for each concept: \\
\textasciigrave\textasciigrave\textasciigrave json\\
{[CONCEPTS]}\\
\textasciigrave\textasciigrave\textasciigrave
\end{prompt}

\begin{prompt}[label={box:llm_judge}]{Orchid}{LLM-as-a-Judge Prompt}
You will be provided with a user query and two responses written by humans or generated by LLMs. \\
Please act as an impartial judge and evaluate the quality of the responses. \\
You should choose the response that follows better the user’s query, is higher in quality, and provides more accurate and relevant information. \\
\{\{\{Begin your evaluation by comparing the two responses. You should think step by step and provide a short explanation.\}\}\}\\
\{\{\{Here are a few examples: {[FEW SHOTS]}\}\}\}\\
Avoid any position biases and ensure that the order in which the responses were presented does not influence your decision. \\
Do not allow the length of the responses to influence your evaluation. \\
Be as objective as possible. \\
\{\{\{After providing your explanation, \}\}\}
Output your final answer, which states which response is better: "A" or "B". \\
Please provide your answers in the JSON format below: \\
\textasciigrave\textasciigrave\textasciigrave json\\
\{\\
\null\quad \{\{\{"explanation": "Your explanation here.",\}\}\}\\
\null\quad "final\_answer": "A" or "B"\\
\}\\
\textasciigrave\textasciigrave\textasciigrave
\end{prompt}

\begin{prompt}[label={box:tie_break}]{Plum}{LLM-as-a-Judge Prompt (Tie Break  -- Concept-guidance)}
You will be provided with a user query and two responses written by humans or generated by LLMs. \\
Please act as an impartial judge and evaluate the quality of the responses. \\
You should choose the response that follows the concepts listed below better. \\
\{\{\{Begin your evaluation by comparing the two responses. You should think step by step and provide a short explanation.\}\}\}\\
Avoid any position biases and ensure that the order in which the responses were presented does not influence your decision. \\
Do not allow the length of the responses to influence your evaluation. \\
Be as objective as possible. \\
\{\{\{After providing your explanation, \}\}\}
Output your final answer, which states which response is better: "A" or "B". \\
Consider *only* the following concepts when evaluating the responses:\\
{[CONCEPT DEFINITIONS]}\\
\\
Please provide your answers in the JSON format below: \\
\textasciigrave\textasciigrave\textasciigrave json\\
\{\\
\null\quad \{\{\{"explanation": "Your explanation here.",\}\}\}\\
\null\quad "final\_answer": "A" or "B"\\
\}\\
\textasciigrave\textasciigrave\textasciigrave
\end{prompt}

\begin{prompt}[label={box:judge_hack_vanilla}]{Gray}{Generating Responses (Judge Hack -- Vanilla)}
You will be provided with a user query, your task is to generate a response to the query. \\
\\
{[USER QUERY]}
\end{prompt}

\begin{prompt}[label={box:judge_hack_concepts}]{YellowOrange}{Generating Responses (Judge Hack -- Concept-guidance)}
You will be provided with a user query, your task is to generate a response to the query. \\
Please consider the following concepts when generating the response:\\
{[CONCEPT DEFINITIONS]}\\
\\
{[USER QUERY]}
\end{prompt}

\end{document}